\documentclass[10pt,twocolumn,letterpaper]{article}

\usepackage[pagenumbers]{cvpr} 

\definecolor{cvprblue}{rgb}{0.21,0.49,0.74}
\usepackage[pagebackref,breaklinks,colorlinks,allcolors=cvprblue]{hyperref}
\usepackage{booktabs}
\usepackage{colortbl}
\usepackage{xcolor}
\def\paperID{}
\def\confName{CVPR}
\def\confYear{2026}

\title{RAP: Fast Feedforward Rendering-Free Attribute-Guided Primitive Importance Score Prediction for Efficient 3D Gaussian Splatting Processing}

\author{
Kaifa Yang$^{1}$ \quad
Qi Yang$^{2*}$ \quad
Yiling Xu$^{1*}$ \quad
Zhu Li$^{2}$\\
$^{1}$Shanghai Jiao Tong University \quad
$^{2}$University of Missouri--Kansas City\\
{\tt\small \{sekiroyyy, yl.xu\}@sjtu.edu.cn \quad
\{qiyang, lizhu\}@umkc.edu}\\
{\small *Corresponding authors}
}

\begin{document}
\maketitle
\begin{abstract}
3D Gaussian Splatting (3DGS) has emerged as a leading technology for high-quality 3D scene reconstruction. However, the iterative refinement and densification process leads to the generation of a large number of primitives, each contributing to the reconstruction to a substantially different extent. Estimating primitive importance is thus crucial, both for removing redundancy during reconstruction and for enabling efficient compression and transmission.
Existing methods typically rely on rendering-based analyses, where each primitive is evaluated through its contribution across multiple camera viewpoints. However, such methods are 1) sensitive to the number and selection of views; 2) rely on specialized differentiable rasterizers; and 3) have long calculation times that grow linearly with view count, making them difficult to integrate as plug-and-play modules, as well as resulting in limited scalability and generalization.
To address these issues, we propose RAP — a fast feedforward Rendering-free Attribute-guided method for efficient importance score Prediction in 3DGS. RAP infers primitive significance directly from intrinsic Gaussian attributes and local neighborhood statistics, avoiding any rendering-based or visibility-dependent computations. A compact MLP is trained to predict per-primitive importance scores using a combination of rendering loss, pruning-aware loss, and significance distribution regularization loss. After being trained on a small set of scenes, RAP generalizes effectively to unseen data and can be seamlessly integrated into reconstruction, compression, and transmission pipelines, providing a unified and efficient pruning solution. Our code is publicly available at: \url{https://github.com/yyyykf/RAP}
\end{abstract}    
\section{Introduction}
\label{sec:intro}

3D Gaussian Splatting (3DGS)~\cite{3dgs} has recently emerged as a powerful explicit representation for novel view synthesis, achieving high-fidelity reconstruction with real-time rendering efficiency. Despite its strengths, 3DGS typically relies on millions of Gaussian primitives to achieve high-fidelity rendering, which imposes a substantial burden on storage and memory. However, these primitives contribute to rendering quality in a highly unbalanced manner: only a fraction of these primitives is valuable for rendering quality, while a large portion is redundant due to either the sub-optimal densification process or incomplete training. 
Therefore, the primitive importance scores are proposed and employed to select and prioritize primitives for differentiated processing, e.g., pruning~\cite{lightgaussian, pup}, compression~\cite{scaffold,hacpp,contextgs,maskgaussian,pcgs,yang2024benchmark}, level-of-detail rendering, and transmission~\cite{FLoD,lapisgs, progs}. For example, prior works such as LightGaussian~\cite{lightgaussian} and PUP-3DGS~\cite{pup} demonstrate that pruning guided by importance scores can significantly reduce the number of primitives without sacrificing fidelity. Consequently, it is crucial to propose an importance estimation method that is accurate, robust, and plug-and-play, serving as a modular component in multiple practical applications of 3DGS.

Existing efforts on primitive importance estimation can be broadly categorized into three groups: attribute-based, rendering-based, and learning-based methods. 1) Attribute-based heuristics~\cite{3dgs, adc} adopt simple rules, such as splitting large primitives during optimization or discarding primitives with opacity below a threshold. While lightweight and independent of camera viewpoints, these strategies ignore the complex blending interactions among overlapping primitives and fail to reflect the true contributions to rendering quality. 
2) Rendering-based approaches \cite{lightgaussian,mesongs,c3dgs,pup} estimate primitive significance by projecting Gaussians onto multiple views to measure their 2D footprints or by evaluating perturbation sensitivity through reconstruction‐error gradients.
These methods offer higher accuracy but are inherently view-dependent, sensitive to the number and choice of views, and computationally expensive, as the time cost grows linearly with view count. Moreover, they require specialized rasterization, limiting their modularity and scalability. 
3) Learning-based approaches \cite{hac,hacpp,contextgs,maskgaussian} jointly optimize a learnable score mask alongside scene reconstruction. Although they can implicitly learn primitive significance, the scores are tightly coupled with specific reconstruction frameworks and thus lack universality. Moreover, once the scene is modified, the precomputed scores become invalid and cannot be reused.
These limitations motivate us to revisit intrinsic attributes as potential signals of primitive importance. Unlike view-dependent or framework-specific strategies, such indicators are lightweight, naturally modular, and generalizable, though their reliability requires further exploration.

\begin{figure*}[htbp]
  \centering
  \includegraphics[width=0.8\linewidth]{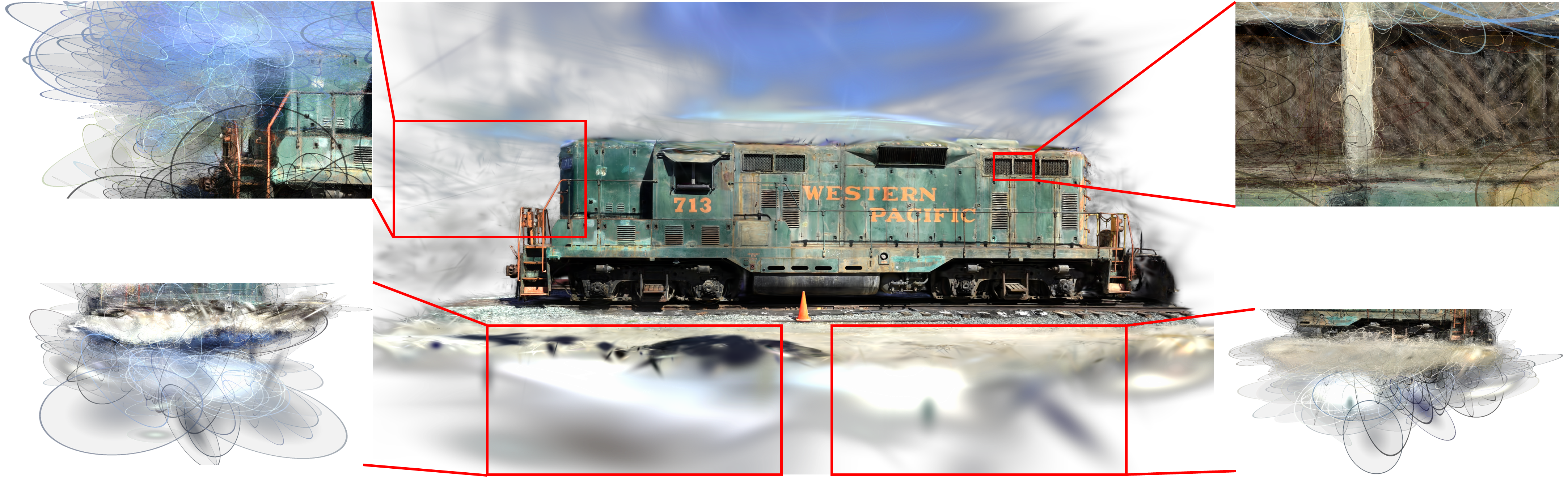}
  \caption{
  Visualization of redundant and low-importance Gaussians in the \textit{Train} scene from Tanks\&Temples. 
  }
  \label{fig:gs_redundancy}
  \vspace{-1.5em}
\end{figure*}

Our key observation, illustrated in Fig.~\ref{fig:gs_redundancy}, is that redundant primitives often exhibit abnormal attributes. For instance, primitives with extremely small scales or very low opacity usually contribute little to rendering quality. Isolated spatial outliers tend to be visually insignificant, though in some cases they may correspond to valid background primitives. Similarly, densification can spawn invisible points with inconsistent colors or near-zero SH coefficients, as they receive little gradient during optimization.
However, no single attribute provides a reliable criterion for importance estimation, as each captures only a limited aspect of information. 
Building on this insight, we propose \textbf{RAP}, a \textbf{R}endering-free \textbf{A}ttribute-guided primitive importance score \textbf{P}rediction framework for estimating primitive importance directly from intrinsic attributes and local neighborhood statistics. Unlike rendering-based or learning-based joint optimization methods, RAP does not require view-dependent computations or per-scene retraining, and also achieves faster calculation with a feedforward style.

Specifically, we extract a 15-dimensional per-point feature vector that captures both global absolute values and local relative statistics, including distance, color, scale, volume, and opacity. 
To model the interactions among these features, we adopt a lightweight MLP that predicts an importance score for each primitive. The network is trained under the supervision of three complementary losses designed to balance fidelity, compactness, and stability: 1) a rendering loss enforces that the pruned model preserves visual fidelity with respect to the ground truth views; 2) A pruning-aware loss prevents the network from trivially assigning high importance to all primitives. This is achieved by regularizing the average predicted score toward a small predefined target, thereby encouraging the network to discard as many primitives as possible and naturally counteracting the rendering loss; 3) A significance distribution regularization ensures that the predicted scores are well separated and distributed from low to high, making downstream tasks, such as pruning decisions, more stable and flexible.
RAP is trained once on a small set of scenes and generalizes well to unseen datasets. Its attribute-based design allows fast, feedforward, plug-and-play importance score prediction without additional scene-specific optimization.  Rendering is only required during the network training, after training, rendering is not necessary for score prediction.

Our contributions are summarized as follows:
\begin{itemize}
    \item We propose RAP, a rendering-free attribute-guided primitive importance score prediction framework that estimates primitive importance directly from intrinsic attributes and local neighborhood statistics.  
    \item We design a set of importance-related per-point features, including average KNN distance, color anisotropy, etc., which together form a compact and discriminative representation of primitive characteristics.  A unified learning framework that based on a lightweight MLP with three tailored loss functions is proposed. These losses jointly constrain the model training, guiding it to produce stable and separable importance score distributions that are friendly to downstream tasks.  
    \item We conduct extensive experiments across diverse datasets and tasks, showing that RAP generalizes well to unseen scenes and consistently improves multiple downstream performance.  
\end{itemize}
\section{Related work}
\label{sec:related}

\subsection{Primitive Importance Score Prediction}
\label{sec:related_method}
Existing approaches for estimating Gaussian primitive importance can be broadly categorized into attribute-based, rendering-based, and learning-based methods.

\textbf{Attribute-based methods} adopt simple heuristics such as Gaussian volume and opacity, which were used in 3DGS~\cite{3dgs} and ADC-GS~\cite{adc} to detect redundant points. These methods are computationally efficient and easy to integrate, but they fail to reliably capture a Gaussian’s actual contribution to rendering quality, especially when multiple Gaussians overlap or interact in complex ways.  

\textbf{Rendering-based methods} derive importance by measuring each Gaussian’s contribution to rendered images, either through visibility or via reconstruction error sensitivity.
LightGaussian~\cite{lightgaussian} computes importance as the 2D projected area multiplied by absolute opacity and further refined by the 3D Gaussian volume. 
PRoGS~\cite{progs} and EAGLES~\cite{eagles} replace absolute opacity with blending opacity obtained during rasterization, which better reflects accumulated visibility but lacks volume refinement. 
MesonGS~\cite{mesongs} and HGSC~\cite{HGSC} combine blending opacity with volume, effectively integrating the strengths of both strategies for more stable scoring.
These strategies provide a closer approximation to perceptual importance but come with notable limitations. 
They rely on custom Gaussian rasterizers, and their computation time increase proportionally with the number of images used for score calculation. 
In addition, their performance is sensitive to the number and distribution of views, as shown in the Supplementary Material (Sec.~\ref{sup:render_views}).
Gradient-based variants estimate significance from reconstruction error sensitivity. 
C3DGS~\cite{c3dgs} uses backward gradients, selecting the maximum SH coefficient channel for color and the maximum eigenvalue of the rotation or scaling matrix for geometry, while PUP-3DGS~\cite{pup} extends this idea by computing perturbation sensitivity from the Hessian of reconstruction errors. 
While gradient signals can effectively indicate importance in well-optimized regions, they often produce unstable scores in poorly reconstructed areas, where large gradients arise from fitting errors rather than true contribution.

\textbf{Learning-based approaches} incorporate joint optimization of a per-Gaussian mask into the reconstruction pipeline, enabling data-driven selection of effective primitives \cite{compact,hac,hacpp,contextgs,maskgaussian}.
The mask implicitly encodes point significance, but these methods are tightly coupled with the reconstruction process, require lengthy optimization, and lack reusability. 
Moreover, once the Gaussian set changes (e.g., after pruning or editing~\cite{gaussianeditor, grouping,vc_edit}), the learned importance distribution becomes invalid, limiting generalization and transferability.

Motivated by these limitations, we propose RAP, a lightweight and rendering-free framework for efficient and generalizable Gaussian importance estimation.

\subsection{Applications of Primitive Importance Score}
\label{sec:app}
The concept of Gaussian primitive importance has been extensively explored in 3D reconstruction, where importance-guided strategies are employed to eliminate redundancy while preserving scene fidelity.
Prune-and-refine approaches \cite{lightgaussian,pup,hybridgs} typically remove low-importance primitives before refining the remaining set, while LP-3DGS \cite{lp3dgs} further learns to determine a favorable pruning ratio.

Alternatively, Mini-Splatting~\cite{minisplatting} selects representative Gaussians through sampling instead of pruning, mitigating potential artifacts from aggressive removal.
More recently, TAMING-3DGS~\cite{taming3dgs} utilizes importance scores to drive targeted densification, enhancing reconstruction efficiency under hardware and resource constraints.

Gaussian primitive importance has also become a key factor in compression and coding frameworks. 
Threshold-based pruning, adopted in MesonGS~\cite{mesongs} and EAGLES~\cite{eagles}, discards low-importance Gaussians prior to encoding to reduce data size. 
Beyond simple thresholding, C3DGS~\cite{c3dgs} employs sensitivity-aware clustering to group Gaussians according to their visual contribution, thereby preserving perceptually important structures while improving compression efficiency. 
Another direction focuses on learnable masking, with methods \cite{hac,hacpp,contextgs,maskgaussian} jointly optimizing mask values during reconstruction to guide the selection of effective primitives.
This idea, first introduced in Compact-3DGS~\cite{compact}, has  been extended in compression-oriented frameworks to enable adaptive importance modeling.

\section{Preliminary}
\label{sec:preliminary}

To design an attribute-guided pruning framework, it is essential to understand how intrinsic Gaussian parameters relate to their perceptual and structural significance. Each Gaussian primitive $\mathcal{G}_i = \{x, y, z, \text{SH}_{0\text{-}47}, o, s_0, s_1, s_2, r_0, r_1, r_2, r_3\}$ encodes its spatial position, view-dependent color, opacity, scale, and rotation.  We observe that the relative importance of Gaussians can be largely inferred from their attributes and local neighborhood relations. 

Four observations reveal the primitive importance: (1) Existing rendering-based methods already provide evidence that opacity and volume are central to significance estimation: by weighting projected areas with these factors, they approximate each Gaussian’s contribution to blending and occlusion. Consistent with this, primitives with \textbf{small scales or low opacity} generally have limited visual influence compared to dominant neighbors, as shown in the top-right region of Fig.~\ref{fig:gs_redundancy}.
(2) Spatial isolation provides another cue for importance estimation. Gaussians are typically aligned with the underlying geometry, while those with \textbf{abnormally large distances to their $k$ nearest neighbors} tend to have lower importance, as illustrated by the floating points in the top-left region of Fig.~\ref{fig:gs_redundancy}.
(3) Under-optimization offers the third evidence for importance prediction. During densification, only Gaussians that remain consistently visible to the training cameras receive stable updates, while those with insufficient visibility fail to converge. Such Gaussians often exhibit an \textbf{abnormal appearance}: some display inconsistent or random colors compared with their neighbors, as seen in the clutter beneath the train (bottom-left and bottom-right of Fig.~\ref{fig:gs_redundancy}); others appear as nearly uniform single-color blobs because their higher-order spherical harmonics remain close to zero, leaving only the low-order terms to dominate the color. Both cases indicate insufficient optimization and, thus, low perceptual importance.
(4) Finally, both global and local magnitude statistics provide complementary signals for importance estimation—\textbf{local relative magnitudes capture occlusion and redundancy, while global magnitudes balance visibility across the scene}.

These observations motivate the construction of a compact feature representation that embeds intrinsic Gaussian attributes with local neighborhood statistics.
Building on this representation, we model the relationship between these features and Gaussian importance. Since explicitly hand-crafting this mapping is challenging, we adopt a lightweight MLP to learn it automatically, as described in Sec.~\ref{sec:method}.
\section{Proposed Method}
\label{sec:method}

\begin{figure*}[t]
  \centering
   \includegraphics[width=0.9\linewidth]{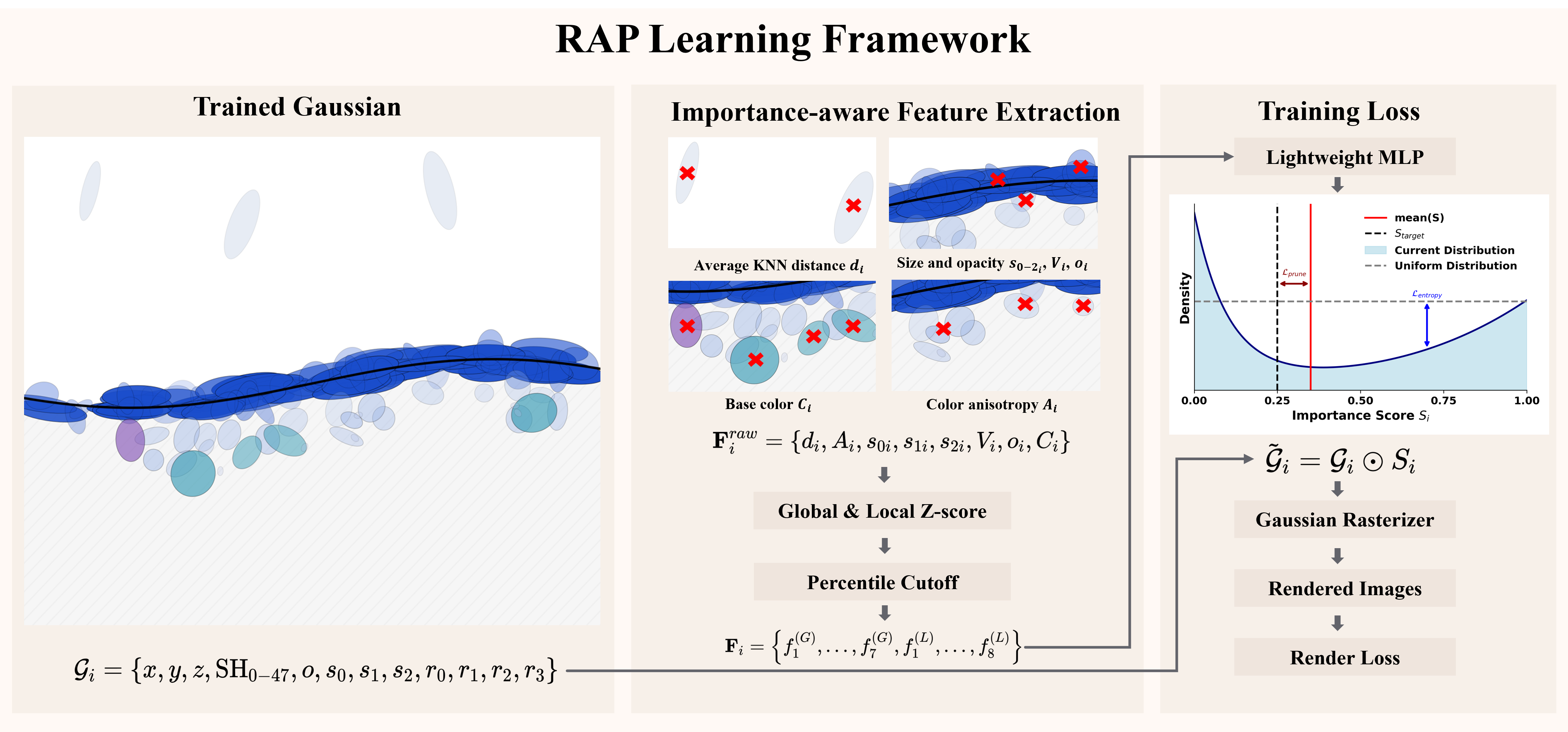}

   \caption{RAP learning framework.}
   \label{fig:framework}
   \vspace{-1.5em}
\end{figure*}

Our proposed RAP framework consists of two stages. First, we perform importance-aware feature extraction, where each Gaussian is represented by a set of importance-related features derived from intrinsic attributes and local neighborhood statistics. These features are then normalized to ensure consistency across different scenes. 
Second, we employ a learning-based importance prediction module, in which a lightweight MLP maps the extracted features to importance scores. Three carefully designed loss functions are proposed to make sure the predicted importance scores are stably distributed from 0 to 1, resulting in flexible, effective, and strong generalization across datasets and downstream GS processing tasks.

\subsection{Importance-aware Feature Extraction}
\label{sec:feature}
We construct a compact 15-dimensional feature vector for each Gaussian primitive by combining intrinsic geometric and appearance attributes with normalized statistics, which consist of three steps: feature computation, feature normalization, and feature concatenation.

$\bullet$\textbf{Feature computation.}
For a Gaussian set $\mathcal{G} = \{\mathcal{G}_1, \ldots, \mathcal{G}_N\}$, each primitive $\mathcal{G}_i$ is represented by a set of raw features,
\begin{equation}
\label{equ:feature}
\mathbf{F}^{raw}_i = \{ d_i, A_i, {s_0}_i, {s_1}_i, {s_2}_i, V_i, o_i, C_i \} \in R^{1\times8},
\end{equation}
where $d_i$ is the average $K$-NN distance, $A_i$ is the color anisotropy, $s_{0,i},s_{1,i},s_{2,i}$ are the sorted Gaussian scales, $V_i$ is the Gaussian volume, $o_i$ is the opacity, and $C_i$ is the DC color. These quantities are computed as follows. 
\begin{itemize}
    \item Average $K$-NN distance:  
    Spatial isolation is measured as
    \begin{equation}
    d_i = \frac{1}{K} \sum_{j \in \mathcal{N}_K(i)} \| \mathbf{p}_i - \mathbf{p}_j \|_2,
    \end{equation}
    where $\mathbf{p}_i \in \mathbb{R}^3$ denotes the spatial position of Gaussian $\mathcal{G}_i$, and $\mathcal{N}_K(i)$ is the set of its $K$ nearest neighbors in Euclidean space.

    \item Color anisotropy:  
    To capture view-dependent appearance variation, we randomly sample $M$ directions $\mathbf{v}_m$ and compute the corresponding RGB colors $\mathbf{c}_i(\mathbf{v}_m)$.  
    The anisotropy is defined as the channel-wise standard deviation across directions:
    \begin{equation}
    A_i = \frac{1}{3} \sum_{c \in \{R,G,B\}} \sigma_m \big( c_i^c(\mathbf{v}_m) \big),
    \end{equation}
    where $\sigma_m$ denotes the standard deviation over sampled directions. A higher $A_i$ indicates a stronger view-dependent color change.
    
    \item Scales and volume:  
    The scales are sorted to ensure rotation invariance:
   $
    s_{0,i} \leq s_{1,i} \leq s_{2,i}.
    $
    The Gaussian volume is then computed as
    \begin{equation}
    V_i = s_{0,i} \times s_{1,i} \times s_{2,i}.
    \end{equation}

    \item Opacity and DC color:  
    Opacity $o_i$ reflects the blending contribution.  
    The DC color $C_i$ is derived from the zeroth-order SH coefficients and averaged over the three RGB channels.
\end{itemize}

$\bullet$\textbf{Feature Normalization.}  
To eliminate scale bias and enhance comparability across Gaussians, each raw feature $f_i \in \mathbf{F}^{raw}_i$ is standardized using both global and local statistics. 
Global normalization $f^{(G)}_i$ applies a scene-wide $z$-score to provide a consistent reference across the entire scene,
where $\mu^{(G)}$ and $\sigma^{(G)}$ denote the global mean and standard deviation across all Gaussians. Local normalization $f^{(L)}_i$ further computes a $K$-NN based $z$-score to emphasize local contrast:
\begin{equation}
f^{(G)}_i = \frac{f_i - \mu^{(G)}}{\sigma^{(G)}}, \quad f^{(L)}_i = \frac{f_i - \mu^{(L)}_i}{\sigma^{(L)}_i},
\end{equation}
where $\mu^{(L)}_i$ and $\sigma^{(L)}_i$ are the local mean and standard deviation over the neighbor set $\mathcal{N}_K(i)$. 
For the DC color $C_i$, only local normalization is applied, as color deviation is typically a consequence of local densification while global normalization would be less meaningful due to natural scene characteristics. 

Since z-score normalization centers features around zero with unit variance, the majority of values are expected to fall within $[-3,3]$ according to the empirical three-sigma rule. In practice, however, extreme outliers may still occur. To enhance robustness, all features are clipped to a fixed percentile range and subsequently linearly rescaled to $[0,1]$.

$\bullet$\textbf{Feature Concatenation.}  
Each Gaussian $\mathcal{G}_i$ is ultimately represented as
\begin{equation}
\mathbf{F}_i = \{f^{(G)}_1, \ldots, f^{(G)}_7, f^{(L)}_1, \ldots, f^{(L)}_8\},
\end{equation}
yielding a 15-dimensional vector (7 global + 8 local) that compactly encodes both intrinsic attributes and normalized neighborhood statistics. This representation provides a discriminative yet lightweight basis for importance prediction.

\subsection{Learning Framework and Optimization}
\label{sec:loss}
The features introduced in Sec.~\ref{sec:feature} capture diverse cues correlated with Gaussian primitive importance, but their interactions are highly nonlinear and difficult to model with handcrafted rules. To address this, we employ a lightweight MLP that learns to map the 15-dimensional feature vector $\mathbf{F}_i$ to an importance score. The network takes the normalized feature vector $\mathbf{F}_i$ as input and produces a single-dimensional output, which is passed through a sigmoid activation to yield an importance score $S_i \in [0,1]$. The architecture is intentionally kept lightweight, avoiding convolutions and graph operations, which enables scalability to millions of Gaussians and ensures efficient inference. 

Given a GS scene, we expect the importance scores to be smoothly distributed from 0 to 1. The importance score of each primitive can be explicitly quantified by the influence of pruning this primitive from the scene. Therefore, we adopt three complementary loss functions, ensuring stable and robust predictions by simulating pruning during the training:

$\bullet$\textbf{Rendering Loss.} The first is a rendering loss, which ensures that the rendering quality after pruning based on importance scores is as high as possible. 
For differentiability, each Gaussian’s opacity and scales are softly reweighted by its predicted importance score $S_i$:
\begin{equation}
\tilde{o}_i = o_i S_i, \quad 
\tilde{\mathbf{s}}_i = \mathbf{s}_i S_i, 
\quad \tilde{\mathcal{G}} = \{\tilde{o}_i, \tilde{\mathbf{s}}_i, \text{others}\}.
\end{equation}
The modified set $\tilde{\mathcal{G}}$ is then rendered through the differentiable rasterizer $\mathcal{R}(\cdot)$.
The loss adopts the standard 3DGS formulation:
\begin{align}
\mathcal{L}_{\text{render}} 
&= (1 - \lambda_{\text{dssim}})
   \mathcal{L}_{\text{1}}(\mathcal{R}(\tilde{\mathcal{G}}), I_{\text{gt}}) \nonumber \\
&\quad + \lambda_{\text{dssim}} \,
   \mathcal{L}_{\text{D-SSIM}}(\mathcal{R}(\tilde{\mathcal{G}}), I_{\text{gt}}).
\end{align}
Here, $I_{\text{gt}}$ is the ground-truth reference image. 
This objective encourages the network to suppress redundant primitives while maintaining perceptual consistency with the ground-truth views.

$\bullet$\textbf{Pruning-Aware Loss.} The second loss is a pruning-aware loss, designed to prevent trivial solutions. 
Without additional constraints and only rendering loss, the network could simply assign high importance to all Gaussians. 
To address this, we regularize the mean predicted score by penalizing deviations from a predefined target:
\begin{equation}
\mathcal{L}_{\text{prune}} = \left( \mathrm{mean}(S_i) - S_{\text{target}} \right)^2,
\end{equation}
where $S_{\text{target}}$ specifies a target mean score that explicitly controls the pruning level.
This objective drives the network to suppress redundant primitives in a manner that counterbalances the rendering loss, ensuring that pruning plays a constructive role in model efficiency without sacrificing rendering fidelity.

$\bullet$\textbf{Distribution Regularization.} The third component is a distribution regularization that encourages smooth and diverse importance scores. 
Entropy acts as an intrinsic measure of prediction dispersion; by maximizing entropy, the model avoids collapsing into trivial binary outcomes, thus enabling flexible and adaptive pruning under varying thresholds.
To approximate entropy in a differentiable manner, we construct a soft histogram with $B$ bins using Gaussian kernels, and compute a normalized entropy as
\begin{equation}
\mathrm{EntropyNorm}(S) = 
-\frac{1}{\log B} 
\sum_{k=1}^{B} \tilde{p}_k \log(\tilde{p}_k + \epsilon),
\end{equation}
where $\tilde{p}_k$ denotes the soft bin occupancy. In our case, we expect a larger entropy, the distribution regularization is thus defined as
\begin{equation}
\mathcal{L}_{\text{entropy}} = 1 - \mathrm{EntropyNorm}(S).
\end{equation}
This objective promotes a well-spread distribution of scores within $[0,1]$, enabling pruning thresholds can be flexibly adjusted.

\textbf{Overall Loss.} The overall training objective combines the three components in a weighted sum:
\begin{equation}
\label{eq:loss_total}
\mathcal{L}_{\text{total}} =
\lambda_{\text{render}} \mathcal{L}_{\text{render}} +
\lambda_{\text{prune}} \mathcal{L}_{\text{prune}} +
\lambda_{\text{entropy}} \mathcal{L}_{\text{entropy}}.
\end{equation}
Here, $\lambda_{\text{render}}$, $\lambda_{\text{prune}}$, and $\lambda_{\text{entropy}}$ balance reconstruction fidelity, pruning strength, and score distribution regularization, respectively.

\textbf{Model Training.} 
In each epoch, one view is randomly sampled from one scene to promote generalization across diverse content. 
During training, pruning is simulated in a differentiable manner by softly reweighting Gaussians with their predicted scores, allowing the network to learn how removal influences rendering quality. 
At inference time, no rendering is performed: the normalized feature vector $\mathbf{F}_i$ is fed through the lightweight MLP once to obtain the importance score $S_i$, after which pruning is applied by a fixed threshold or a percentile-based ratio. 
This deployment protocol requires neither scene-specific retraining nor additional rendering, enabling fast, plug-and-play pruning on unseen datasets.

\definecolor{rank1}{RGB}{255, 190, 128}   
\definecolor{rank2}{RGB}{255, 210, 160}   
\definecolor{rank3}{RGB}{255, 230, 190}   
\definecolor{baselinebg}{RGB}{245, 245, 245} 

\newcommand{\rankone}[1]{\cellcolor{rank1}\textbf{#1}}
\newcommand{\ranktwo}[1]{\cellcolor{rank2}\textbf{#1}}
\newcommand{\rankthree}[1]{\cellcolor{rank3}\textbf{#1}}
\newcommand{\baseline}[1]{\cellcolor{baselinebg}#1}

\section{Experiments}
\label{sec:exp}

\subsection{Implementation Details}
\label{sec:exp_imp}

The MLP of RAP consists of three hidden layers with widths of 32, 32, and 16 neurons, respectively.
We set $\lambda_{\text{dssim}} = 0.2$, $\lambda_{\text{render}} = 1.0$, $\lambda_{\text{prune}} = 1.0$, and $\lambda_{\text{entropy}} = 0.25$ for the loss terms. 
Feature extraction uses $K = 128$ nearest neighbors for local statistics, $B = 250$ bins for the entropy approximation, and a Gaussian kernel width of $\sigma = 0.01$. 
The MLP is trained for 15{,}000 iterations on 10 randomly selected scenes from the DL3DV-10K~\cite{dl3dv} dataset.

We evaluate RAP on the same three benchmarks as the original 3DGS~\cite{3dgs}, including the full set of scenes from Mip-NeRF360~\cite{mip360} (five outdoor and four indoor scenes), two scenes from DeepBlending~\cite{db}, and two scenes from Tanks\&Temples~\cite{tandt}. 
All scenes are trained following the standard 3DGS protocol, using $7/8$ of the views for training and the remaining $1/8$ for testing.

We compare RAP with multiple baselines, including a simple opacity-thresholding heuristic, visibility-based approaches such as LightGaussian~\cite{lightgaussian}, MesonGS~\cite{mesongs}, and EAGLES~\cite{eagles}, and gradient-based approaches such as C3DGS~\cite{c3dgs} and PUP-3DGS~\cite{pup}. 
For rendering-based baselines, importance scores are computed using the training views to ensure a fair comparison under identical conditions.

\subsection{Post-hoc Pruning on Trained 3DGS}
\label{sec:exp1_post_pruning}
We perform post-hoc pruning experiments on pretrained 3DGS representations by retaining a fixed percentage (from 5\% to 95\%) of the Gaussians with the highest predicted importance scores. 

\begin{figure}[t]
  \centering
   \includegraphics[width=\linewidth]{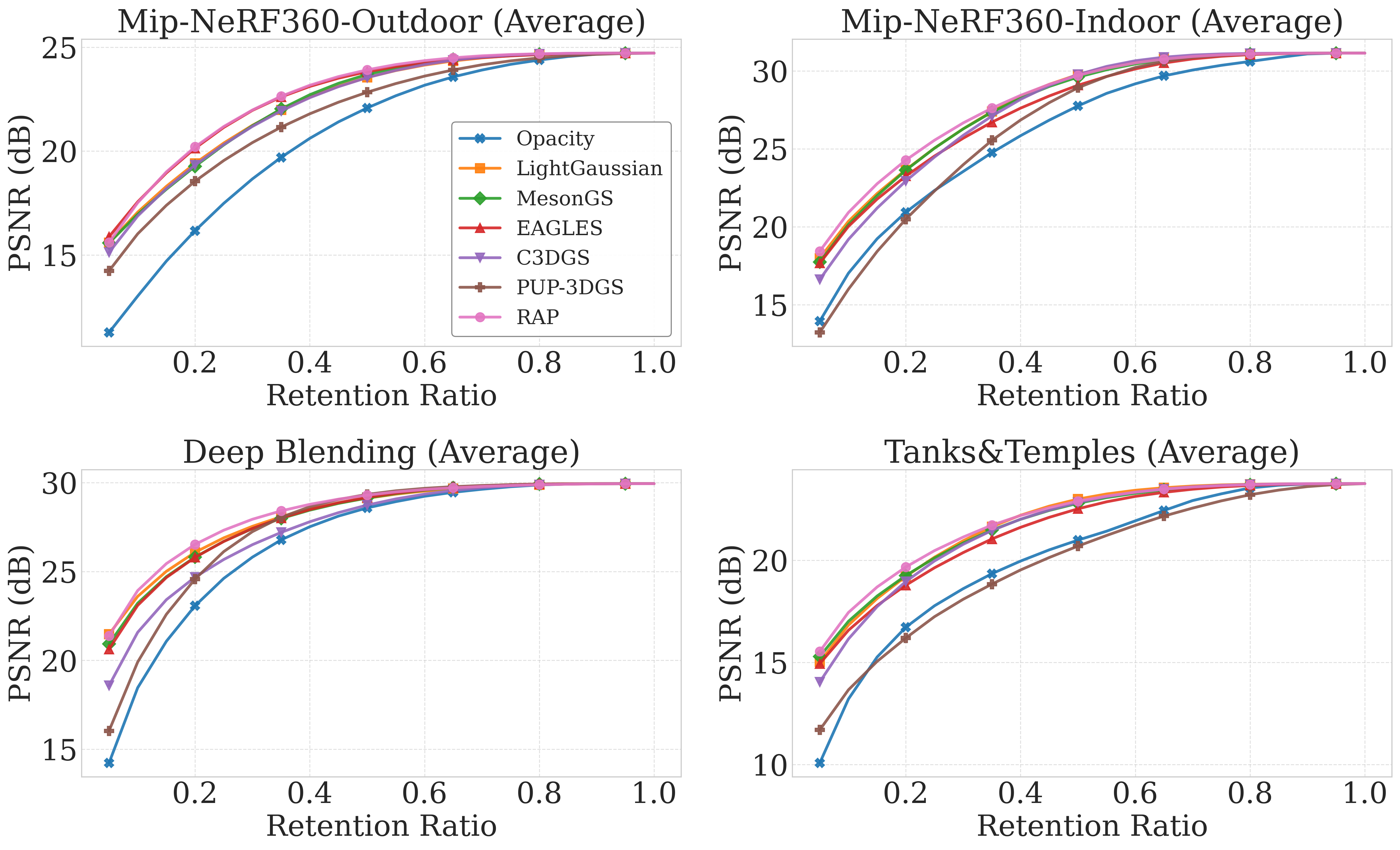}
   \caption{PSNR vs. retention ratio curves. }
   \label{fig:exp1_rd}
   \vspace{-1em}
\end{figure}

Performance is evaluated by plotting retention-ratio vs. reconstruction-quality curves (PSNR, SSIM, LPIPS) in Fig.~\ref{fig:exp1_rd}. 
Due to space constraints, we report only the average PSNR across the Mip-NeRF360-Outdoor, Mip-NeRF360-Indoor, Deep Blending, and Tanks\&Temples datasets in the main paper; per-scene results and the full set of SSIM and LPIPS metrics are provided in the supplementary material. 

Considering that pruning can be used as part of GS compression, to further quantify pruning efficiency, we compute the BD-Rate (Bjøntegaard Delta Bitrate) inspired by compression research of each method relative to the opacity-based baseline, using the average per-scene storage size as the rate axis. 
This metric provides a compact assessment of rate-distortion behavior across datasets---a lower BD-Rate indicates superior reconstruction quality under comparable storage budgets. 
The summarized results are presented in Table~\ref{tab:exp1_bdbr}.

\label{subsec:post_hoc}

\begin{table}[t]
    \centering
    \caption{BD-Rate (\%) comparison of pruning-based Gaussian Splatting methods relative to the opacity baseline. Lower BD-Rate indicates better rate–distortion efficiency}
    \label{tab:exp1_bdbr}
    \resizebox{\columnwidth}{!}{%
    \begin{tabular}{|l|c|c|c|c|c|c|}
    \hline
    \textbf{} & \textbf{LightGS} & \textbf{MesonGS} & \textbf{EAGLES} & \textbf{C3DGS} & \textbf{PUP-3DGS} & \textbf{RAP} \\ \hline
    \textbf{Mip-Outdoor} 
        & \rankthree{-35.21} 
        & -34.89 
        & \ranktwo{-41.28} 
        & -33.77 
        & -22.54 
        & \rankone{-42.63} \\ \hline
    \textbf{Mip-Indoor}  
        & \ranktwo{-31.15} 
        & \rankthree{-30.34} 
        & -24.98 
        & -27.63 
        & -8.70 
        & \rankone{-33.90} \\ \hline
    \textbf{Deep Blending}       
        & \ranktwo{-30.72} 
        & -28.84 
        & \rankthree{-29.87} 
        & -16.28 
        & -19.46 
        & \rankone{-36.76} \\ \hline
    \textbf{Tanks\&Temples}      
        & \ranktwo{-37.98} 
        & \rankthree{-36.12} 
        & -30.01 
        & -34.37 
        & 7.06 
        & \rankone{-40.11} \\ \hline
    \end{tabular}%
    }
    \vspace{-1.5em}
\end{table}
As shown in Fig.~\ref{fig:exp1_rd} and Table~\ref{tab:exp1_bdbr}, our method consistently outperforms prior approaches across all datasets and pruning ratios. 
While the performance gap is relatively small under light pruning (e.g., retention $>$50\%), it becomes increasingly pronounced as pruning becomes more aggressive. 
At a pruning ratio of 60\%, RAP achieves up to a 0.5~dB PSNR gain over competing methods. 
The superiority is further reflected in BD-Rate reductions, with improvements such as $-42.63\%$ on the Mip-NeRF360-Outdoor dataset.

We also observe a notable difference in robustness between visibility-based and gradient-based methods. The former (e.g., LightGaussian, MesonGS, EAGLES) exhibit more consistent performance across datasets, while the latter (e.g., C3DGS, PUP-3DGS) show larger variance. For instance, C3DGS performs comparably on Mip-NeRF360 but suffers over 1 dB drop on Deep Blending. On Tanks\&Temples, PUP-3DGS even underperforms the opacity baseline, with a positive BD-Rate of {+7.06\%}, indicating poor generalization across diverse scenes.

Lastly, we find that when the importance estimation function is sufficiently accurate, up to 40\% of Gaussians can be pruned with negligible degradation in rendering quality. Even at a 60\% pruning rate, the average PSNR drop remains within 2 dB. These results suggest that reliable importance scoring can enable highly compact representations without significantly compromising visual fidelity.

\begin{table}[]
    \centering
    \caption{Importance score computation time (seconds).}
    \label{tab:exp_importance_time}
    \resizebox{\columnwidth}{!}{%
    \begin{tabular}{|l|c|c|c|c|c|c|c|}
    \hline
    \textbf{Dataset} & \textbf{Opacity} & \textbf{LightGS} & \textbf{MesonGS} & \textbf{EAGLES} & \textbf{C3DGS} & \textbf{PUP-3DGS} & \textbf{RAP} \\ \hline
    \textbf{Mip-Outdoor} & \rankone{3.73} & 15.86 & 42.40 & 42.96 & \ranktwo{11.69} & 20.95 & 15.64 \\ \hline
    \textbf{Mip-Indoor}  & \rankone{1.27} & 22.71 & 14.84 & 14.97 & 15.96 & 21.22 & \ranktwo{5.72} \\ \hline
    \textbf{Deep Blending} & \rankone{2.99} & 20.20 & 31.30 & 31.65 & 13.62 & 20.28 & \ranktwo{11.64} \\ \hline
    \textbf{Tanks\&Temples} & \rankone{2.53} & 18.62 & 17.53 & 13.92 & 9.22 & 20.50 & \ranktwo{6.66} \\ \hline
    \end{tabular}%
    }
    \vspace{-1.5em}
\end{table}

Table~\ref{tab:exp_importance_time} reports the computation time for importance score prediction (including point loading and metric computation). 
RAP is among the fastest methods across datasets: it ranks second (behind the trivial opacity baseline) on Mip-NeRF360-Indoor, Deep Blending, and Tanks\&Temples, and third on Mip-NeRF360-Outdoor where C3DGS is slightly faster (11.69\,s vs.\ 15.64\,s for RAP). 
Despite this single exception, RAP consistently outpaces all visibility-based approaches (LightGaussian, MesonGS, EAGLES) by a wide margin, and is generally faster than gradient-based methods (C3DGS, PUP-3DGS). 
This advantage stems from RAP’s feature-driven, rendering-free design, which avoids per-view rasterization and back-propagation; consequently, runtime scales with the number of primitives rather than with the number of training views, enabling efficient deployment across diverse scenes.

\subsection{Pruning-in-the-Loop Training}
\label{sec:exp2_rec_pruning}
To evaluate the potential of integrating pruning directly into the 3DGS generation process, we modify the pipeline to periodically remove redundant Gaussians during optimization. This experiment is motivated by the observation in Sec.~\ref{subsec:post_hoc}, where pruning up to 40\% of Gaussians post-training leads to negligible quality degradation.

Specifically, we embed the pruning strategy into the densification phase: every \textbf{1500} training iterations, we apply a pruning operation to remove the bottom \textbf{40\%} of Gaussians ranked by the importance score predicted by each method are removed. To assess the effectiveness of this integration, we compare the final reconstructions against the standard 3DGS training baseline (without pruning) using four metrics: PSNR, SSIM, LPIPS, and storage size (MB).

\begin{table*}[hbpt]
\centering
\caption{Reconstruction quality and size comparison under integrated pruning (40\% pruning every 1500 iterations).}
\label{tab:exp_reconstruction}
\resizebox{0.9\textwidth}{!}{%
\begin{tabular}{lcccccccccccccccc}
\toprule
\textbf{Method} & \multicolumn{4}{c}{\textbf{MipNeRF360 Outdoor}} & \multicolumn{4}{c}{\textbf{MipNeRF360 Indoor}} & \multicolumn{4}{c}{\textbf{Deep Blending}} & \multicolumn{4}{c}{\textbf{Tanks \& Temples}} \\
\cmidrule(lr){2-5} \cmidrule(lr){6-9} \cmidrule(lr){10-13} \cmidrule(lr){14-17}
& PSNR ↑ & SSIM ↑ & LPIPS ↓ & Size ↓ 
& PSNR ↑ & SSIM ↑ & LPIPS ↓ & Size ↓ 
& PSNR ↑ & SSIM ↑ & LPIPS ↓ & Size ↓ 
& PSNR ↑ & SSIM ↑ & LPIPS ↓ & Size ↓ \\
\midrule
\rowcolor{baselinebg}
3DGS & \baseline{24.688} & \baseline{0.729} & \baseline{0.238} & \baseline{926.895} 
     & \baseline{31.076} & \baseline{0.925} & \baseline{0.186} & \baseline{299.202} 
     & \baseline{29.817} & \baseline{0.907} & \baseline{0.239} & \baseline{586.064} 
     & \baseline{23.734} & \baseline{0.853} & \baseline{0.169} & \baseline{372.036} \\
\midrule
Opacity & 24.550 & 0.718 & 0.275 & 313.657 
        & \rankthree{30.604} & \rankthree{0.912} & 0.216 & 72.176 
        & \ranktwo{29.899} & \rankone{0.910} & \rankone{0.248} & 149.873 
        & \ranktwo{23.674} & \ranktwo{0.842} & \rankthree{0.200} & 118.414 \\

LightGaussian & \ranktwo{24.676} & \rankone{0.728} & \rankone{0.255} & 300.907 
        & \rankone{30.777} & \rankone{0.917} & \rankone{0.205} & 71.676 
        & \rankthree{29.839} & \rankthree{0.908} & \ranktwo{0.249} & 155.207 
        & \rankthree{23.634} & \rankone{0.845} & \rankone{0.191} & 118.048 \\

MesonGS & 24.554 & \rankthree{0.723} & \ranktwo{0.256} & 297.255 
        & 30.173 & \ranktwo{0.914} & \ranktwo{0.207} & 71.587 
        & 29.638 & 0.904 & 0.254 & 154.137 
        & 23.541 & \rankthree{0.839} & \ranktwo{0.197} & 115.922 \\

EAGLES & 24.520 & 0.722 & \ranktwo{0.256} & \rankthree{295.899} 
       & 29.801 & 0.911 & \rankthree{0.210} & 71.356 
       & 29.551 & 0.904 & 0.254 & 154.352 
       & 23.212 & 0.834 & 0.202 & \rankthree{113.500} \\

C3DGS & \rankthree{24.638} & \rankthree{0.723} & 0.262 & 303.820 
      & 30.271 & \rankthree{0.912} & 0.212 & \ranktwo{67.821} 
      & 29.739 & 0.905 & 0.258 & \rankone{139.431} 
      & 23.421 & 0.836 & 0.205 & 115.941 \\

PUP-3DGS & 24.328 & 0.714 & \rankthree{0.261} & \rankone{281.821} 
         & 29.422 & 0.909 & 0.212 & \rankthree{70.056} 
         & 29.699 & \rankthree{0.908} & \rankone{0.248} & \ranktwo{145.294} 
         & 22.365 & 0.816 & 0.215 & \rankone{107.662} \\

RAP & \rankone{24.709} & \ranktwo{0.726} & 0.263 & \ranktwo{285.067} 
    & \ranktwo{30.696} & 0.911 & 0.218 & \rankone{67.804} 
    & \rankone{29.913} & \ranktwo{0.909} & \rankthree{0.252} & \rankthree{147.250} 
    & \rankone{23.774} & \ranktwo{0.842} & 0.201 & \ranktwo{113.236} \\
\bottomrule
\end{tabular}
}
\vspace{-1.5em}
\end{table*}

The results, summarized in Table~\ref{tab:exp_reconstruction}, show that integrating pruning introduces negligible quality degradation while achieving substantial model size reduction. The reconstructed Gaussian sets are only about one-third to one-fifth the size of the original models, with similar performance across methods. The proposed RAP reports higher PSNRs than vanilla GS on MipNeRF360 Outdoor, Deep Blending, and Tanks \& Temples, while other approaches generally give a PSNR drop of 0.3--0.5~dB. It indicates that the proposed RAP can facilitate the optimization process via accurately removing redundant primitives, as well as resulting in a better convergence direction. We found that all the pruning methods demonstrate a PSNR decrease on MipNeRF360 Indoor, and an obviously larger drop (around 1.5~dB) is observed with PUP-3DGS and EAGLES, revealing that this dataset is more challenging than the others.   

On the other hand, directly using opacity as the pruning score also yields competitive PSNR results---ranking third on MipNeRF360-Outdoor and second on Deep~Blending and Tanks\&Temples. This can be explained by the densification behavior: when pruning is sub-optimal, the densification process tends to over-generate new Gaussians to compensate, resulting in more primitives but still competitive reconstruction quality, which can be found by the model size of ``Opacity".

\subsection{Integration with MPEG GSC}

\begin{figure}[t]
  \centering
  \begin{subfigure}{0.48\linewidth}
    \centering
    \includegraphics[width=\linewidth]{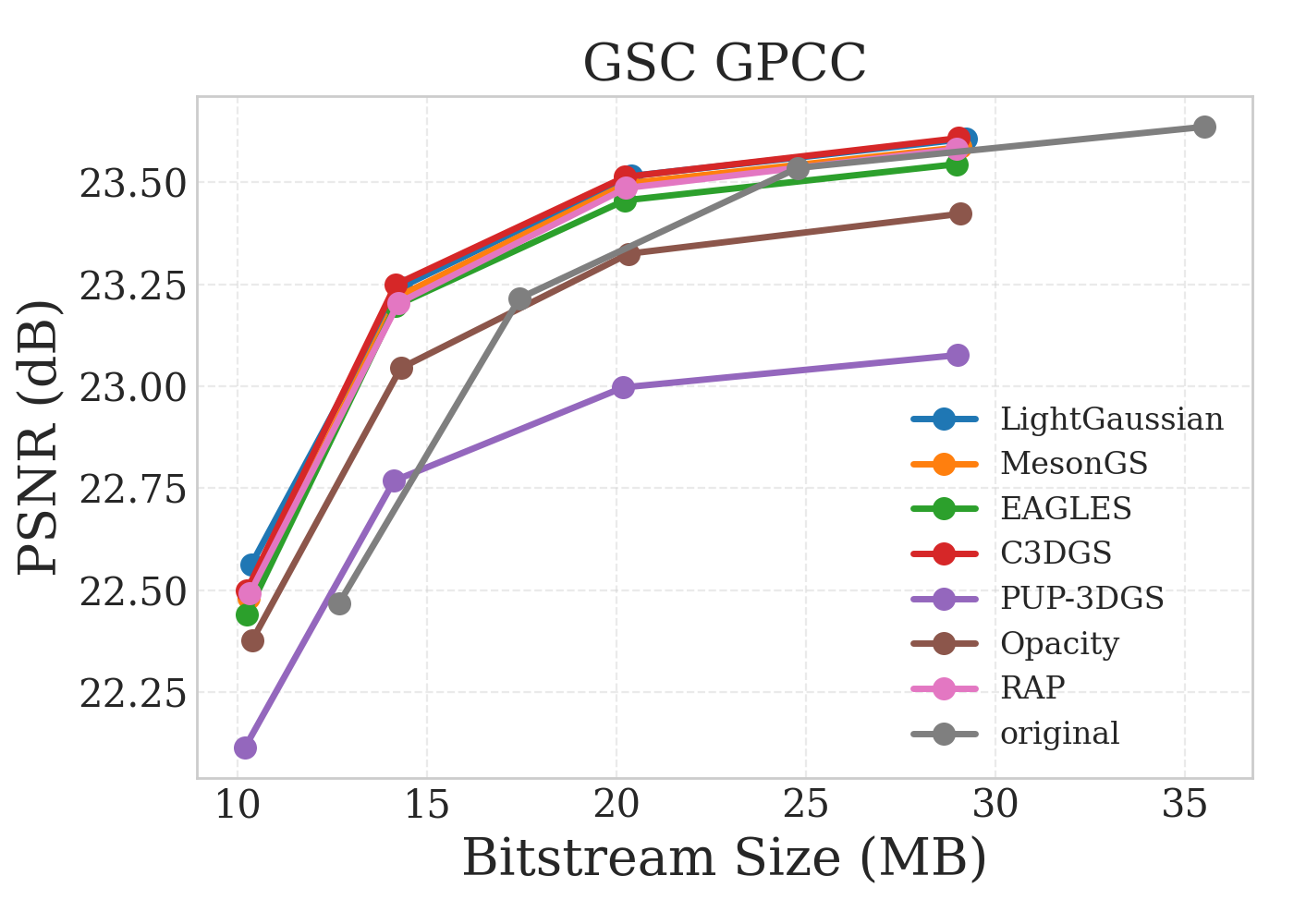}
    \caption{GSC GPCC}
    \label{fig:gsc_gpcc_avg}
  \end{subfigure}
  \hfill
  \begin{subfigure}{0.48\linewidth}
    \centering
    \includegraphics[width=\linewidth]{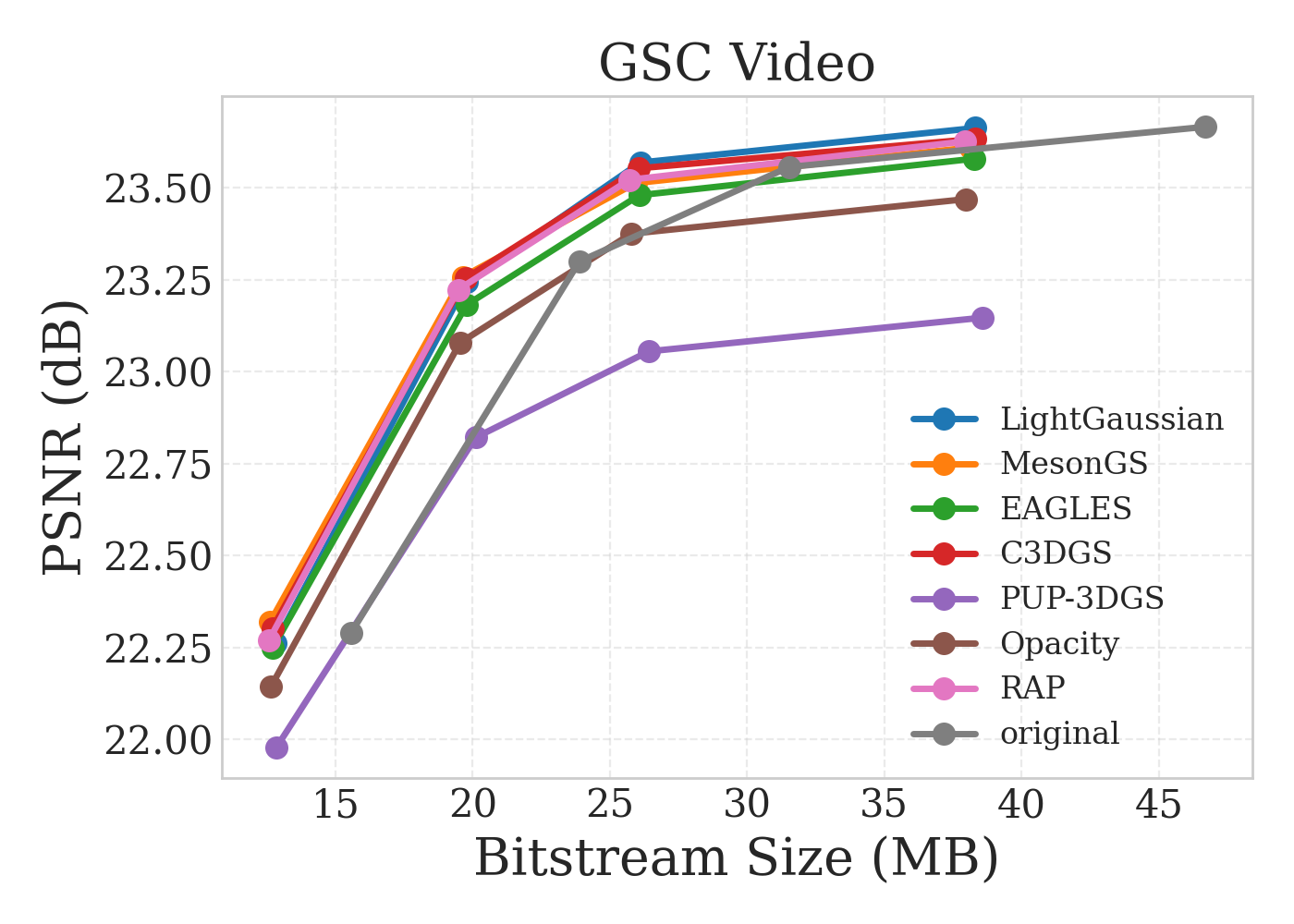}
    \caption{GSC Video}
    \label{fig:gsc_video_avg}
  \end{subfigure}

  \caption{
    R-D curves for GSC with 20\% pruning
  }
  \label{fig:gsc_avg}
  \vspace{-1.5em}
\end{figure}
To evaluate whether importance-guided pruning benefits downstream compression, 
we integrate the pruning step as a pre-processing module into both branches of 
MPEG Gaussian Splat Coding (GSC)~\cite{mpeg_gsc}: the G-PCC–based~\cite{gpcc} 
and the video-based pipeline~\cite{gscodec_studio}. 
For each scene, 20\% of the Gaussians are removed according to the predicted importance  scores, and the remaining primitives are encoded using identical codec settings at four 
rate points. Rate–distortion (R-D) curves are reported in Fig.~\ref{fig:gsc_avg}. Importance-guided pruning consistently improves coding efficiency for both branches. Most pruning methods achieve \textbf{15--20\%} BD-Rate gains.
RAP remains in the top performance band, indicating strong generalization under diverse scene conditions.
In contrast, PUP-3DGS and simple opacity thresholding exhibit unstable behavior across datasets. 
 
Although the G-PCC--based pipeline attains higher PSNR at lower bitrates than the video-based one, both benefit substantially from pre-pruning. 
Overall, these results show that effective importance-guided pruning produces compact yet encoder-friendly Gaussian sets and significantly boosts GSC coding efficiency regardless of the codec design.

\subsection{Ablation Studies}
\label{sec:abla}
\begin{figure}[t]
  \centering

  \begin{subfigure}{0.48\linewidth}
    \centering
    \includegraphics[width=\linewidth]{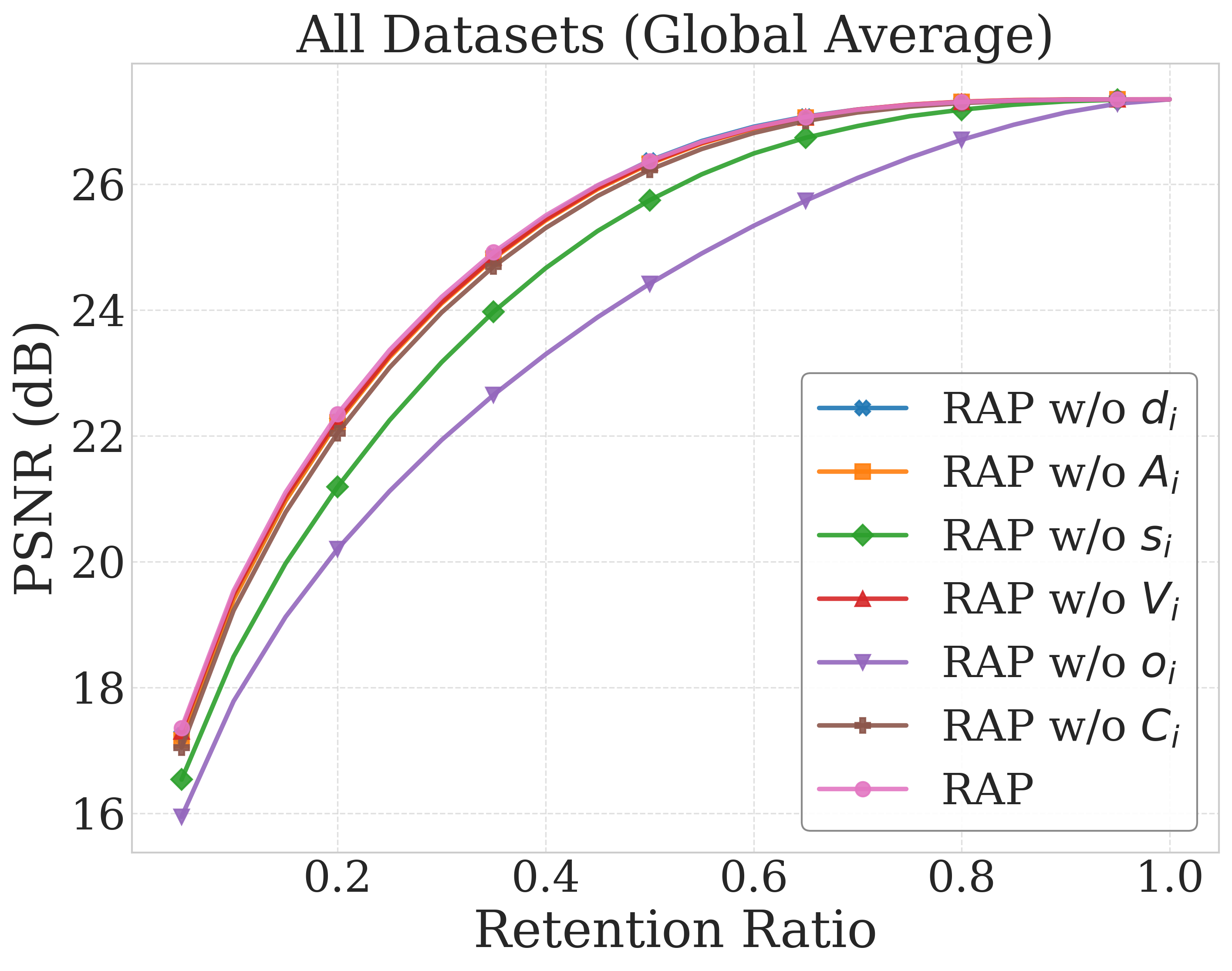}
    \caption{Individual features}
    \label{fig:abla_feat1}
  \end{subfigure}
  \hfill
  \begin{subfigure}{0.48\linewidth}
    \centering
    \includegraphics[width=\linewidth]{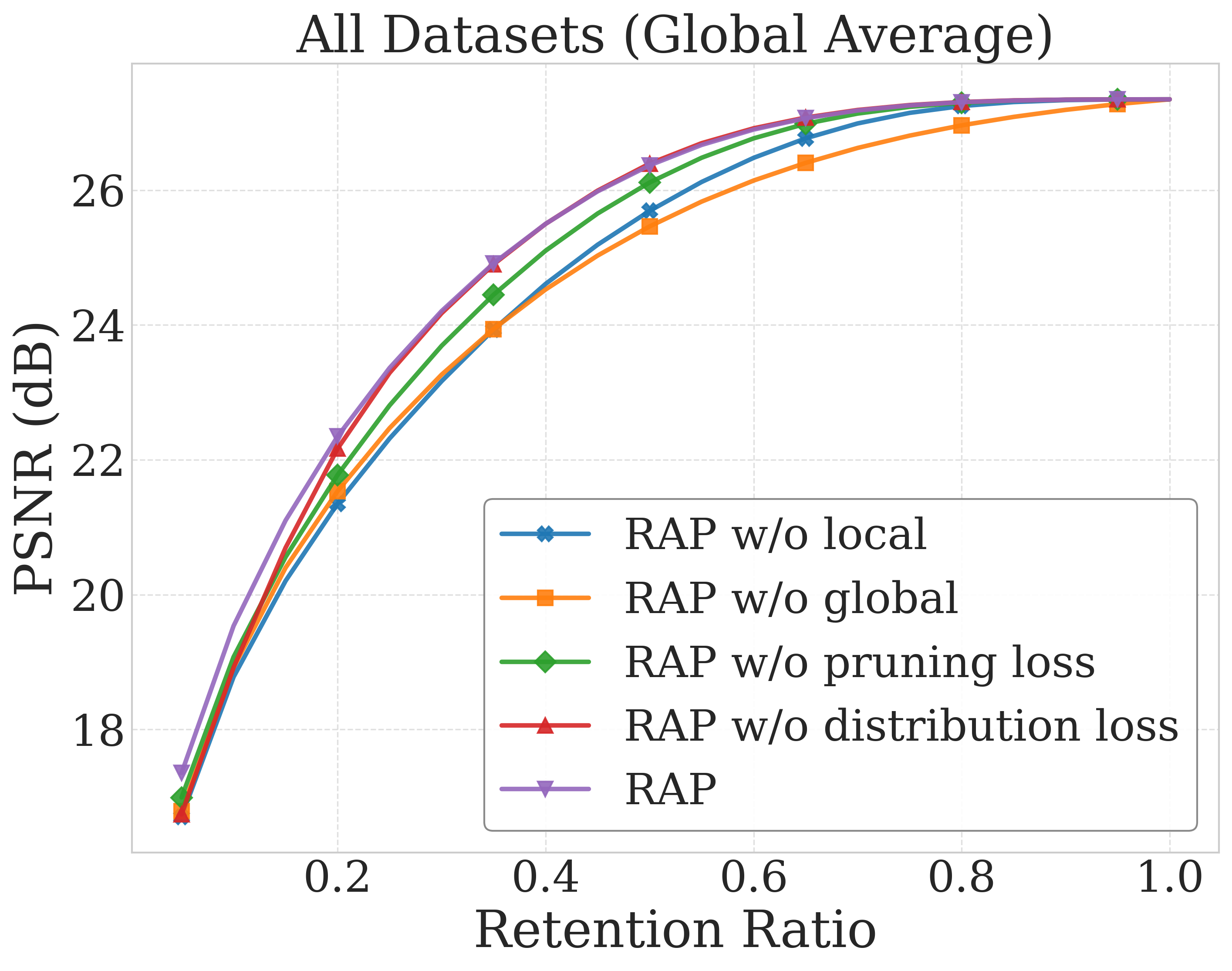}
    \caption{Normalization and loss}
    \label{fig:abla_feat2}
  \end{subfigure}

  \caption{
    Ablation studies of the proposed feature and loss.
  }
  \label{fig:abla_half}
  \vspace{-1.5em}
\end{figure}
To better understand the behavior and design choices of RAP, we conduct comprehensive ablation studies feature effectiveness and loss formulation.
We perform a series of experiments to analyze two aspects: 
(1) the effectiveness of the proposed feature design and normalization strategy; and 
(2) the contribution of each loss term in our RAP training.

$\bullet$\textbf{Effectiveness of feature design.}
As formulated in Equation~\ref{equ:feature}, RAP integrates geometric and appearance descriptors 
$\{d_i, A_i, s_{0,i}, s_{1,i}, s_{2,i}, V_i, o_i, C_i\}$ 
together with their local and global normalized counterparts.
To assess the contribution of each component, we perform two ablation experiments.

Fig.~\ref{fig:abla_feat1} evaluates the contribution of each feature by removing one attribute at a time while keeping all others unchanged. Opacity $o_i$ proves to be the most crucial cue—its removal leads to about 1-2~dB PSNR drop at the same pruning ratio. 
Gaussian scales $\{s_{0,i}, s_{1,i}, s_{2,i}\}$ follow as the next most influential features, causing around 0.5–1~dB degradation when excluded. 
Other features, including color anisotropy $A_i$ and average KNN distance $d_i$, provide smaller yet consistent benefits, particularly at aggressive pruning ratios, indicating their complementary role. 

$\bullet$\textbf{Normalization and loss formulation.}
Fig.~\ref{fig:abla_feat2} analyzes the effect of normalization and loss design.  
Removing either local or global normalization leads to a substantial PSNR drop (1.5--2~dB), confirming their complementary roles: 
local normalization enhances intra-region contrast to better separate redundant primitives, 
while global normalization ensures scene-level consistency and stabilizes feature magnitudes across datasets. 
Together, they support robust and generalizable importance estimation under diverse scene conditions.

The pruning-aware loss also plays a critical role. 
Removing it results in a moderate but consistent degradation (around 0.5~dB on average), as the network loses explicit control over the expected pruning ratio. 
As visualized in the supplementary material (Sec.~\ref{sup:loss}), the absence of this loss causes the predicted score distribution to shift toward a higher mean, reflecting overly conservative pruning.

The distribution loss contributes smaller yet steady improvements, with a noticeable gain at low retention ratios (approximately 0.25~dB). 
This effect is most evident in dense and complex scenes, where enforcing a smooth score distribution helps differentiate subtle importance variations among overlapping primitives. 
Without this loss, the predicted scores collapse toward near-binary outputs around 0 and~1 (see Sec.~\ref{sup:loss}), reducing the model's flexibility to support arbitrary pruning ratios.

\section{Conclusion and Limitations}
\label{sec:conclusion}
We presented RAP, a fast feedforward rendering-free and attribute-guided framework for estimating primitive importance in 3DGS. By leveraging intrinsic attributes and normalized neighborhood statistics, RAP avoids view-dependent rendering analysis while achieving strong generalization, efficient inference, and consistent improvements in pruning, reconstruction, and compression. Our current integration with downstream tasks relies on pruning a fixed global ratio based on predicted scores. A more principled coupling with reconstruction and compression remains open—particularly how to allocate different sampling densities across regions or how to enable hierarchical or region-adaptive coding within GSC. We leave these directions for future exploration.
{
    \small
    \bibliographystyle{ieeenat_fullname}
    \bibliography{main}

@String(CVPR= {IEEE Conf. Comput. Vis. Pattern Recog.})

@String(ECCV= {Eur. Conf. Comput. Vis.})

@String(NIPS= {Adv. Neural Inform. Process. Syst.})

@String(TOG= {ACM Trans. Graph.})

@String(ACMMM= {ACM Int. Conf. Multimedia})

@String(ICASSP=	{ICASSP})

@String(IJCAI = {IJCAI})

@String(CVPR  = {CVPR})

@String(ECCV  = {ECCV})

@String(NIPS  = {NeurIPS})

@String(TOG   = {ACM TOG})

@String(ACMMM = {ACM MM})

@article{lightgaussian,
  title={Lightgaussian: Unbounded 3D Gaussian Compression with 15x Reduction and 200+ FPS},
  author={Fan, Zhiwen and Wang, Kevin and Wen, Kairun and Zhu, Zehao and Xu, Dejia and Wang, Zhangyang and others},
  journal={NIPS},
  volume={37},
  pages={140138--140158},
  year={2024}
}

@inproceedings{pup,
  title={PUP 3D-GS: Principled Uncertainty Pruning for 3D Gaussian Splatting},
  author={Hanson, Alex and Tu, Allen and Singla, Vasu and Jayawardhana, Mayuka and Zwicker, Matthias and Goldstein, Tom},
  booktitle={CVPR},
  pages={5949--5958},
  year={2025}
}

@inproceedings{mesongs,
  title={MesonGS: Post-training Compression of 3D Gaussians via Efficient Attribute Transformation},
  author={Xie, Shuzhao and Zhang, Weixiang and Tang, Chen and Bai, Yunpeng and Lu, Rongwei and Ge, Shijia and Wang, Zhi},
  booktitle={ECCV},
  pages={434--452},
  year={2024},
  organization={Springer}
}

@inproceedings{c3dgs,
  title={Compressed 3D Gaussian Splatting for Accelerated Novel View Synthesis},
  author={Niedermayr, Simon and Stumpfegger, Josef and Westermann, R{\"u}diger},
  booktitle={CVPR},
  pages={10349--10358},
  year={2024}
}

@inproceedings{hac,
  title={HAC: Hash-grid Assisted Context for 3D Gaussian Splatting Compression},
  author={Chen, Yihang and Wu, Qianyi and Lin, Weiyao and Harandi, Mehrtash and Cai, Jianfei},
  booktitle={ECCV},
  pages={422--438},
  year={2024},
  organization={Springer}
}

@article{hacpp,
  title={Hac++: Towards 100x compression of 3d gaussian splatting},
  author={Chen, Yihang and Wu, Qianyi and Lin, Weiyao and Harandi, Mehrtash and Cai, Jianfei},
  journal={arXiv preprint arXiv:2501.12255},
  year={2025}
}

@article{contextgs,
  title={ContextGS: Compact 3D Gaussian Splatting with Anchor Level Context Model},
  author={Wang, Yufei and Li, Zhihao and Guo, Lanqing and Yang, Wenhan and Kot, Alex and Wen, Bihan},
  journal={NIPS},
  volume={37},
  pages={51532--51551},
  year={2024}
}

@inproceedings{maskgaussian,
  title={MaskGaussian: Adaptive 3D Gaussian Representation from Probabilistic Masks},
  author={Liu, Yifei and Zhong, Zhihang and Zhan, Yifan and Xu, Sheng and Sun, Xiao},
  booktitle={CVPR},
  pages={681--690},
  year={2025}
}

@inproceedings{taming3dgs,
  title={Taming 3DGS: High-Quality Radiance Fields with Limited Resources},
  author={Mallick, Saswat Subhajyoti and Goel, Rahul and Kerbl, Bernhard and Steinberger, Markus and Carrasco, Francisco Vicente and De La Torre, Fernando},
  booktitle={SIGGRAPH Asia},
  pages={1--11},
  year={2024}
}

@inproceedings{minisplatting,
  title={Mini-splatting: Representing scenes with a constrained number of gaussians},
  author={Fang, Guangchi and Wang, Bing},
  booktitle={ECCV},
  pages={165--181},
  year={2024},
  organization={Springer}
}

@inproceedings{compact,
  title={Compact 3d gaussian representation for radiance field},
  author={Lee, Joo Chan and Rho, Daniel and Sun, Xiangyu and Ko, Jong Hwan and Park, Eunbyung},
  booktitle={CVPR},
  pages={21719--21728},
  year={2024}
}

@inproceedings{eagles,
  title={Eagles: Efficient accelerated 3d gaussians with lightweight encodings},
  author={Girish, Sharath and Gupta, Kamal and Shrivastava, Abhinav},
  booktitle={ECCV},
  pages={54--71},
  year={2024},
  organization={Springer}
}

@inproceedings{dl3dv,
  title={Dl3dv-10k: A large-scale scene dataset for deep learning-based 3d vision},
  author={Ling, Lu and Sheng, Yichen and Tu, Zhi and Zhao, Wentian and Xin, Cheng and Wan, Kun and Yu, Lantao and Guo, Qianyu and Yu, Zixun and Lu, Yawen and others},
  booktitle={CVPR},
  pages={22160--22169},
  year={2024}
}

@inproceedings{mip360,
  title={Mip-nerf 360: Unbounded anti-aliased neural radiance fields},
  author={Barron, Jonathan T and Mildenhall, Ben and Verbin, Dor and Srinivasan, Pratul P and Hedman, Peter},
  booktitle={CVPR},
  pages={5470--5479},
  year={2022}
}

@article{tandt,
  title={Tanks and temples: Benchmarking large-scale scene reconstruction},
  author={Knapitsch, Arno and Park, Jaesik and Zhou, Qian-Yi and Koltun, Vladlen},
  journal={ACM Transactions on Graphics (ToG)},
  volume={36},
  number={4},
  pages={1--13},
  year={2017},
  publisher={ACM New York, NY, USA}
}

@article{db,
  title={Deep blending for free-viewpoint image-based rendering},
  author={Hedman, Peter and Philip, Julien and Price, True and Frahm, Jan-Michael and Drettakis, George and Brostow, Gabriel},
  journal={ACM Transactions on Graphics (ToG)},
  volume={37},
  number={6},
  pages={1--15},
  year={2018},
  publisher={ACM New York, NY, USA}
}

@article{3dgs,
  title={3D Gaussian splatting for real-time radiance field rendering.},
  author={Kerbl, Bernhard and Kopanas, Georgios and Leimk{\"u}hler, Thomas and Drettakis, George},
  journal={ACM Trans. Graph.},
  volume={42},
  number={4},
  pages={139--1},
  year={2023}
}

@misc{mpeg_gsc,
  title        = {Description of JEE 6.3 on exploration of {3DGS} representation and coding technologies},
  author       = {{ISO/IEC JTC 1/SC 29/WG 07}},
  year         = {2025},
}

@techreport{gpcc,
  author      = {{ISO/IEC JTC 1/SC 29}},
  title       = {Information technology — Coded representation of immersive media — Part 9: Geometry-based point cloud compression},
  institution = {ISO/IEC},
  number      = {ISO/IEC 23090-9:2023},
  edition     = {1},
  year        = {2023},
  month       = {Mar},
  url         = {https://www.iso.org/standard/78990.html}
}

@article{gscodec_studio,
  title={GSCodec Studio: A Modular Framework for Gaussian Splat Compression},
  author={Li, Sicheng and Wu, Chengzhen and Li, Hao and Gao, Xiang and Liao, Yiyi and Yu, Lu},
  journal={arXiv preprint arXiv:2506.01822},
  year={2025}
}

@INPROCEEDINGS{adc,
  title={ADC-GS: Anchor-Driven Deformable and Compressed Gaussian Splatting for Dynamic Scene Reconstruction},
  author={Huang, He and Yang, Qi and Liu, Mufan and Xu, Yiling and Li, Zhu},
  booktitle={IJCAI},
  year={2025}
}

@INPROCEEDINGS{HGSC,
  author={Huang, He and Huang, Wenjie and Yang, Qi and Xu, Yiling and Li, Zhu},
  booktitle={ICASSP}, 
  title={A Hierarchical Compression Technique for 3D Gaussian Splatting Compression}, 
  year={2025},
  volume={},
  number={},
  pages={1-5},
  doi={10.1109/ICASSP49660.2025.10887742}
}

@inproceedings{progs,
  title={Progs: Progressive rendering of gaussian splats},
  author={Zoomers, Brent and Wijnants, Maarten and Molenaers, Ivan and Vanherck, Joni and Put, Jeroen and Michiels, Nick},
  booktitle={WACV},
  pages={3118--3127},
  year={2025},
  organization={IEEE}
}

@inproceedings{hybridgs,
  title={HybridGS: High-Efficiency Gaussian Splatting Data Compression using Dual-Channel Sparse Representation and Point Cloud Encoder},
  author={Yang, Qi and Yang, Le and Van Der Auwera, Geert and Li, Zhu},
  booktitle={ICML},
  year={2025}
}

@inproceedings{gaussianeditor,
  title={Gaussianeditor: Swift and controllable 3d editing with gaussian splatting},
  author={Chen, Yiwen and Chen, Zilong and Zhang, Chi and Wang, Feng and Yang, Xiaofeng and Wang, Yikai and Cai, Zhongang and Yang, Lei and Liu, Huaping and Lin, Guosheng},
  booktitle={CVPR},
  pages={21476--21485},
  year={2024}
}

@inproceedings{scaffold,
  title={Scaffold-gs: Structured 3d gaussians for view-adaptive rendering},
  author={Lu, Tao and Yu, Mulin and Xu, Linning and Xiangli, Yuanbo and Wang, Limin and Lin, Dahua and Dai, Bo},
  booktitle={CVPR},
  pages={20654--20664},
  year={2024}
}

@article{FLoD,
    author = {Seo, Yunji and Choi, Young Sun and Son, HyunSeung and Uh, Youngjung},
    title = {FLoD: Integrating Flexible Level of Detail into 3D Gaussian Splatting for Customizable Rendering},
    year = {2025},
    issue_date = {August 2025},
    publisher = {Association for Computing Machinery},
    address = {New York, NY, USA},
    volume = {44},
    number = {4},
    issn = {0730-0301},
    url = {https://doi.org/10.1145/3731430},
    doi = {10.1145/3731430},
    journal = {ACM Trans. Graph.},
    month = jul,
    articleno = {112},
    numpages = {16},
}

@inproceedings{lapisgs,
  title={Lapisgs: Layered progressive 3d gaussian splatting for adaptive streaming},
  author={Shi, Yuang and Morin, G{\'e}raldine and Gasparini, Simone and Ooi, Wei Tsang},
  booktitle={3DV},
  pages={991--1000},
  year={2025},
  organization={IEEE}
}

@article{pcgs,
  title={Pcgs: Progressive compression of 3d gaussian splatting},
  author={Chen, Yihang and Li, Mengyao and Wu, Qianyi and Lin, Weiyao and Harandi, Mehrtash and Cai, Jianfei},
  journal={arXiv preprint arXiv:2503.08511},
  year={2025}
}

@inproceedings{grouping,
  title={Gaussian grouping: Segment and edit anything in 3d scenes},
  author={Ye, Mingqiao and Danelljan, Martin and Yu, Fisher and Ke, Lei},
  booktitle={ECCV},
  pages={162--179},
  year={2024},
  organization={Springer}
}

@inproceedings{vc_edit,
  title={View-consistent 3d editing with gaussian splatting},
  author={Wang, Yuxuan and Yi, Xuanyu and Wu, Zike and Zhao, Na and Chen, Long and Zhang, Hanwang},
  booktitle={ECCV},
  pages={404--420},
  year={2024},
  organization={Springer}
}

@article{lp3dgs,
  title={Lp-3dgs: Learning to prune 3d gaussian splatting},
  author={Zhang, Zhaoliang and Song, Tianchen and Lee, Yongjae and Yang, Li and Peng, Cheng and Chellappa, Rama and Fan, Deliang},
  journal={NIPS},
  volume={37},
  pages={122434--122457},
  year={2024}
}

@inproceedings{yang2024benchmark,
  title={A benchmark for gaussian splatting compression and quality assessment study},
  author={Yang, Qi and Yang, Kaifa and Xing, Yuke and Xu, Yiling and Li, Zhu},
  booktitle={ACMMM Asia},
  pages={1--8},
  year={2024}
}
}

\clearpage
\setcounter{page}{1}
\maketitlesupplementary


\section{Robustness of rendering-based methods.}
\label{sup:render_views}
\begin{figure}[htbp]
  \centering
  \includegraphics[width=\linewidth]{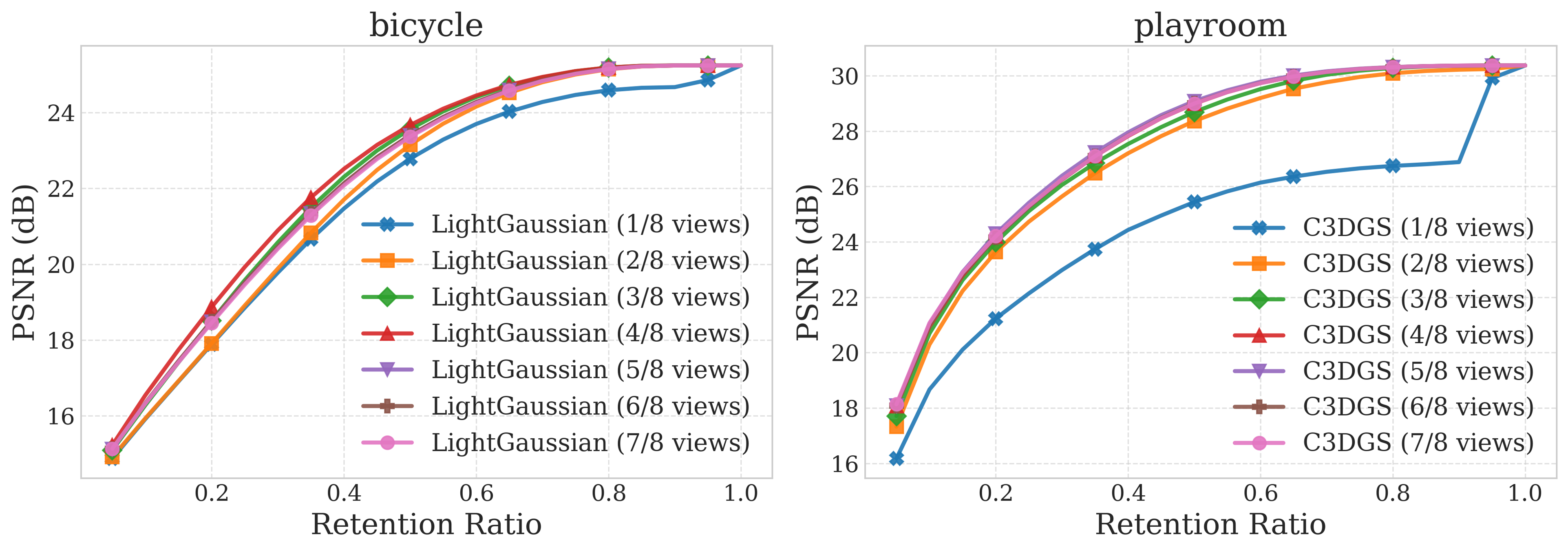}
  \caption{
  Robustness of rendering-based importance estimation to the number of training views used.
  Left: LightGaussian on the \textit{bicycle} scene from Mip-NeRF360-Outdoor. 
  Right: C3DGS on the \textit{playroom} scene from Deep Blending.
  Each curve corresponds to a different subset of views ($1/8$–$7/8$) used for score computation.
  }
  \label{fig:ablation_viewnum}
\end{figure}

As discussed in Section~\ref{sec:exp_imp}, rendering-based baselines estimate importance by projecting Gaussians from all available training views (i.e., $7/8$ of the total views). 
To analyze their sensitivity to the number of views, we uniformly subsample $1/8$, $2/8$, $\dots$, $7/8$ of the training views to recompute importance scores and evaluate post-hoc pruning results.
As shown in Fig.~\ref{fig:ablation_viewnum}, rendering-based methods exhibit notable instability when the number or distribution of input views changes. 
For instance, LightGaussian’s pruning quality on the \textit{bicycle} scene fluctuates by up to 1.5~dB PSNR across different view counts, while C3DGS on \textit{playroom} suffers a 4~dB drop when using only $1/8$ of the views and still varies by about 1~dB across the remaining subsets. 
Besides, using more views does not always yield better scores: LightGaussian achieves its best performance with $4/8$ views, and C3DGS peaks at $5/8$.
These results suggest that rendering-based importance estimation is highly dependent on the number and spatial distribution of selected views, 
whereas our RAP, relying solely on point-level features without rendering, is inherently view-agnostic and thus more robust.

\section{Post-hoc Pruning: Additional Results}
\label{sup:post_hoc}

\begin{figure*}[htpb]
  \centering
  \begin{minipage}{0.48\linewidth}
    \centering
    \includegraphics[width=\linewidth]{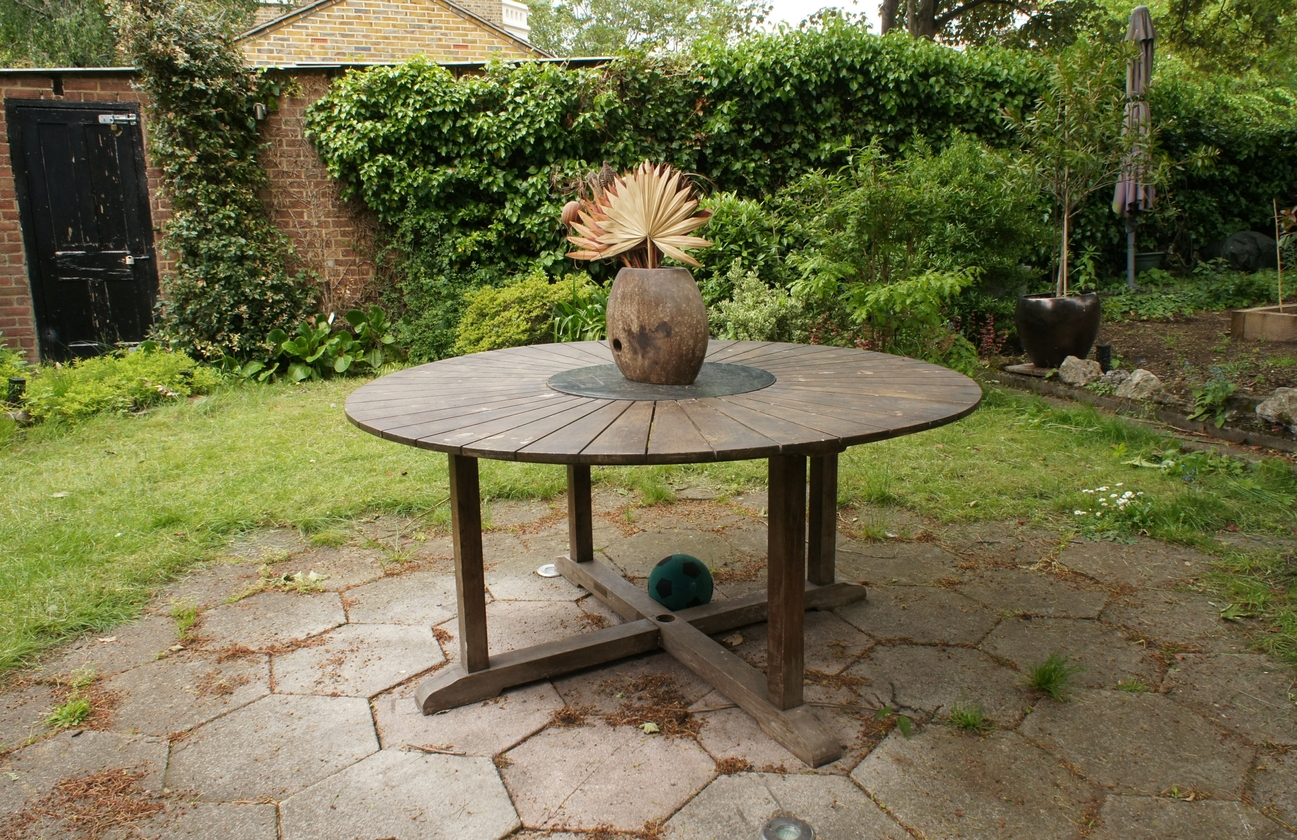}
    \vspace{2pt}
    \small (a) Original scene
  \end{minipage}
  \hfill
  \begin{minipage}{0.48\linewidth}
    \centering
    \includegraphics[width=\linewidth]{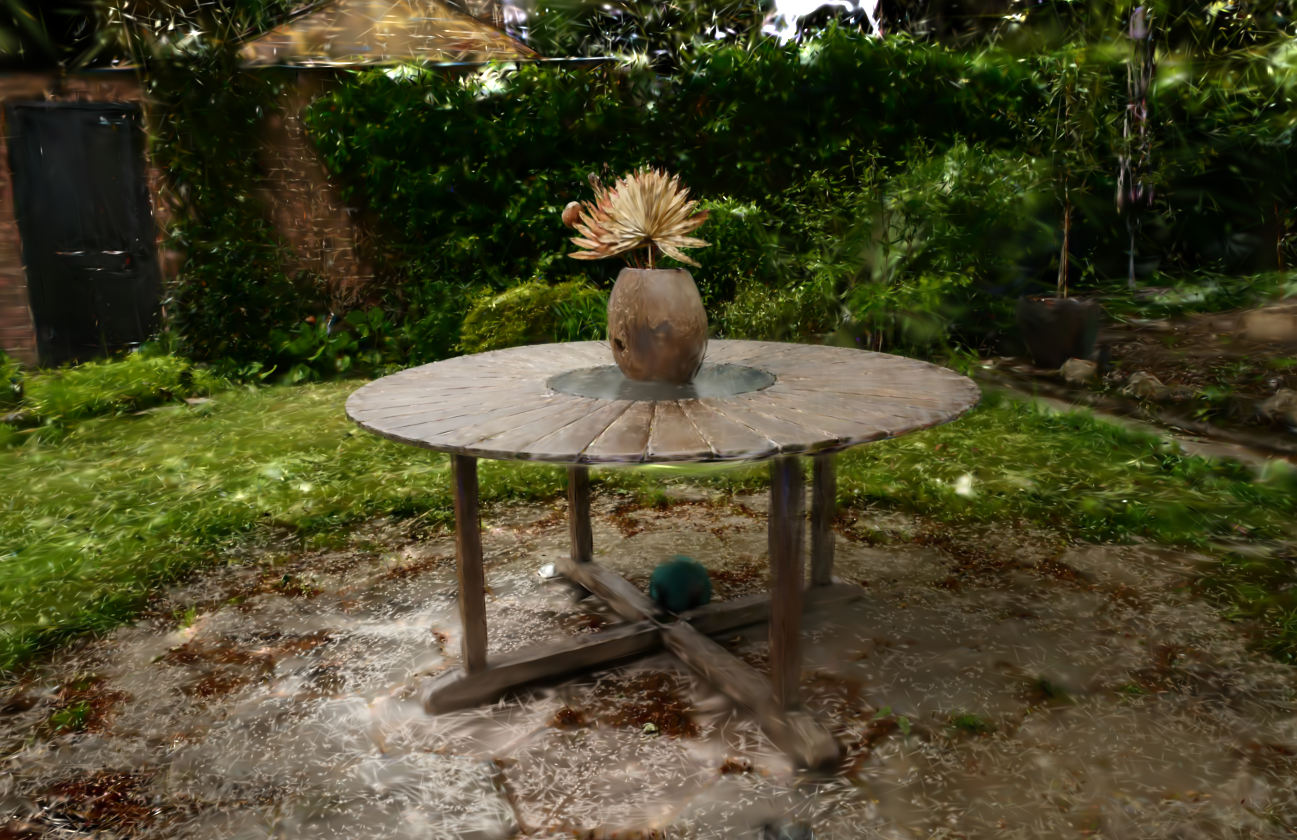}
    \vspace{2pt}
    \small (b) RAP (ours, 5\% retained)
  \end{minipage}
  \vspace{4pt}
  \begin{minipage}{0.48\linewidth}
    \centering
    \includegraphics[width=\linewidth]{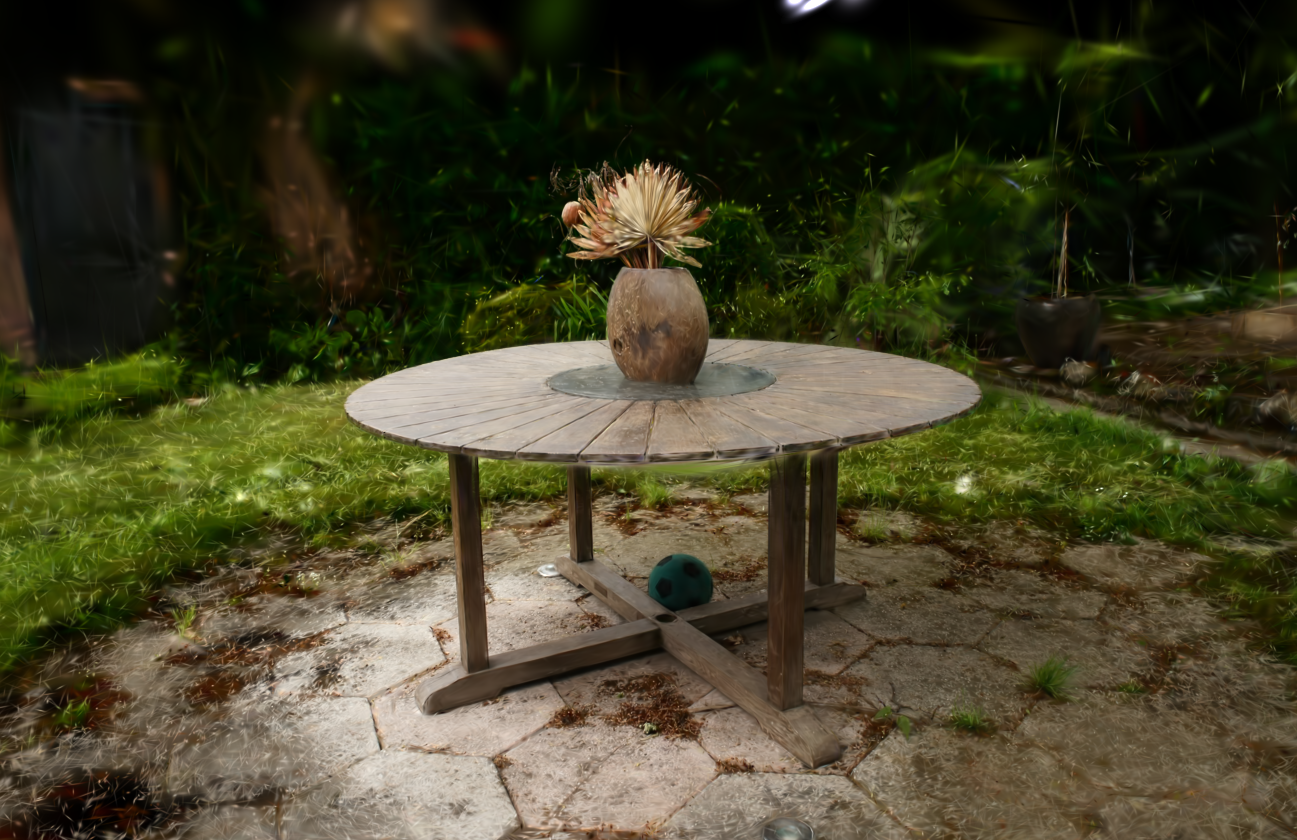}
    \vspace{2pt}
    \small (c) EAGLES (visibility-based, 5\% retained)
  \end{minipage}
  \hfill
  \begin{minipage}{0.48\linewidth}
    \centering
    \includegraphics[width=\linewidth]{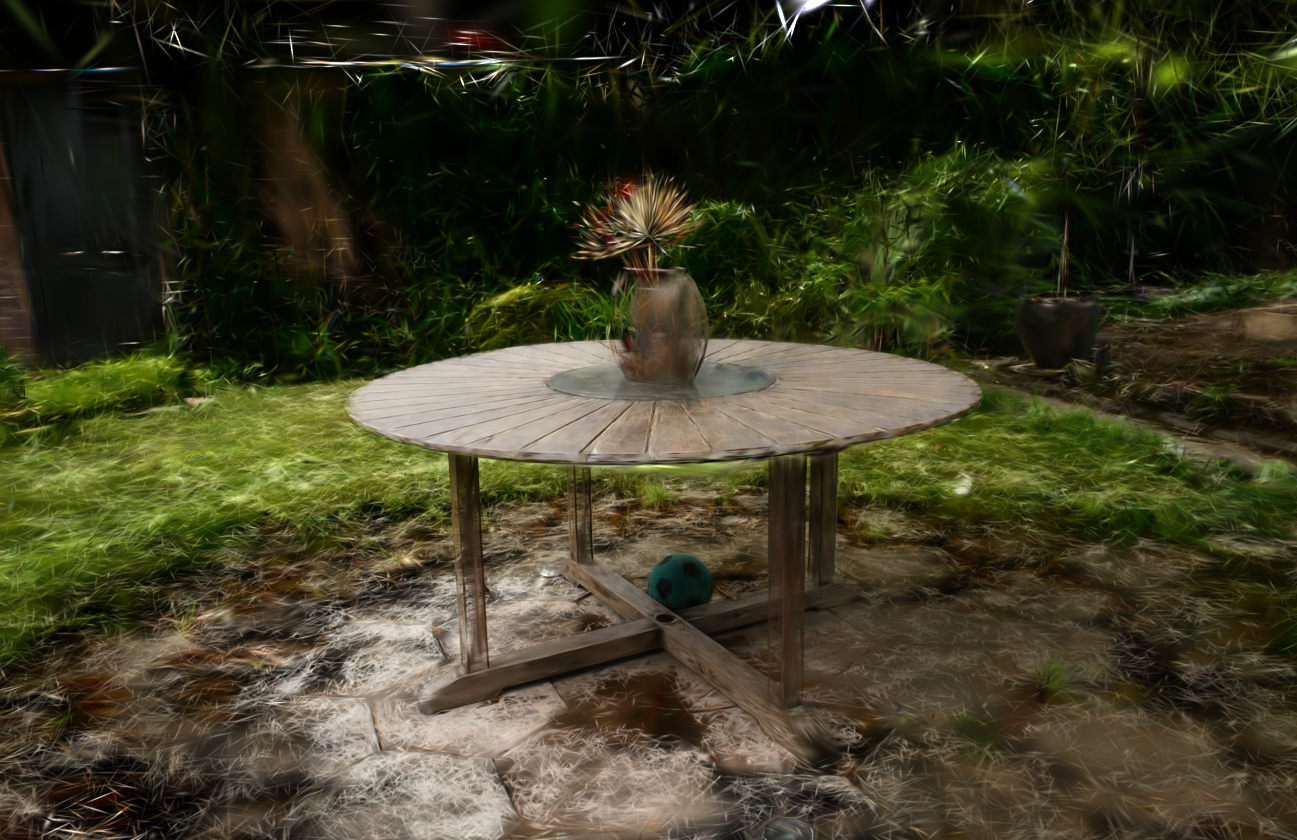}
    \vspace{2pt}
    \small (d) C3DGS (gradient-based, 5\% retained)
  \end{minipage}
  \vspace{4pt}
  \caption{
    Pruning behavior of different importance estimators at 5\% retention. 
    EAGLES favors central regions, C3DGS keeps edges, while RAP produces a more uniform, structure-preserving subset.
  }
  \label{fig:exp_garden_vis}
\end{figure*}

\begin{figure*}[htbp]
  \centering
  \includegraphics[width=\linewidth,height=0.31\textheight,keepaspectratio]{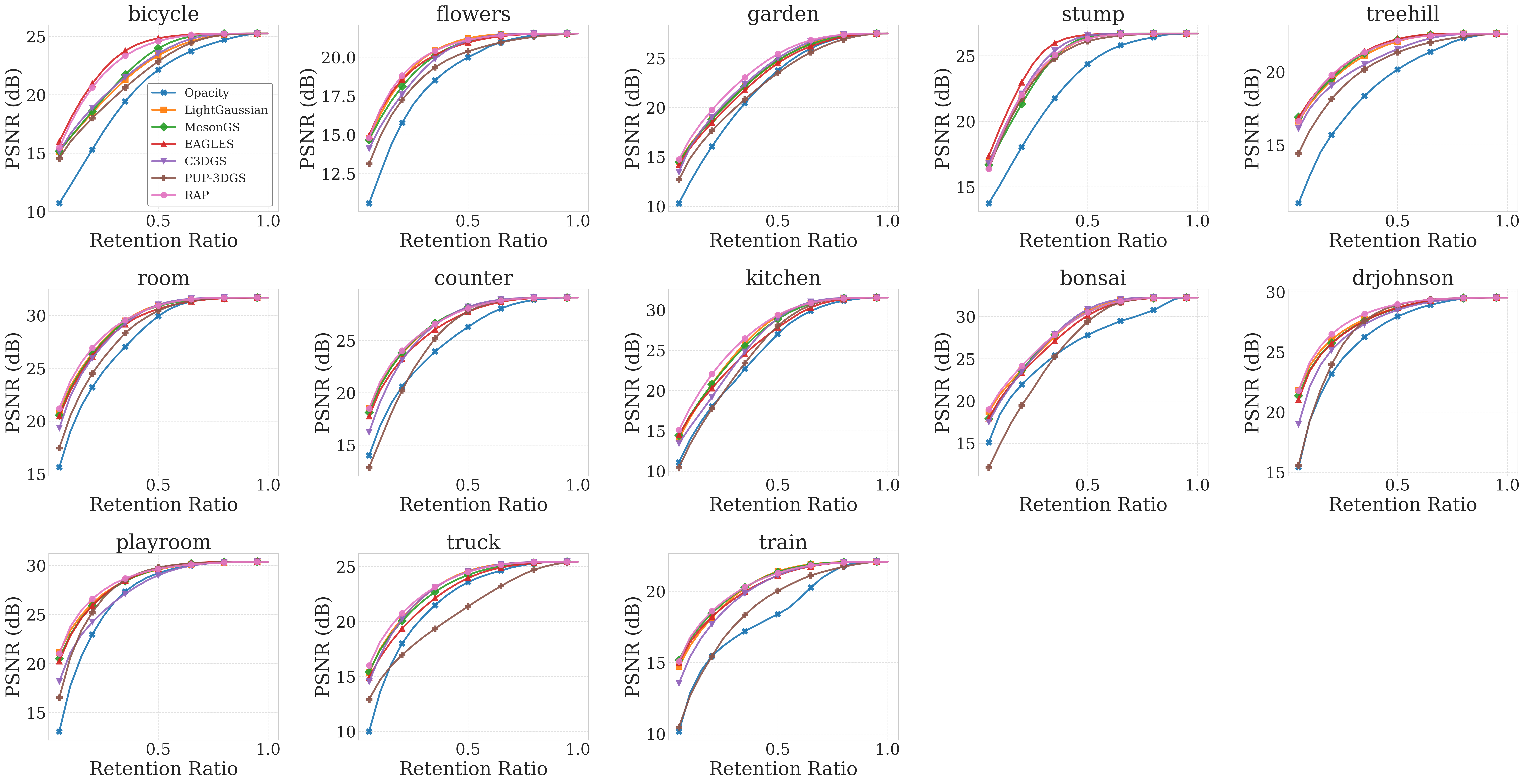}
  \vspace{-3pt}
  \includegraphics[width=\linewidth,height=0.31\textheight,keepaspectratio]{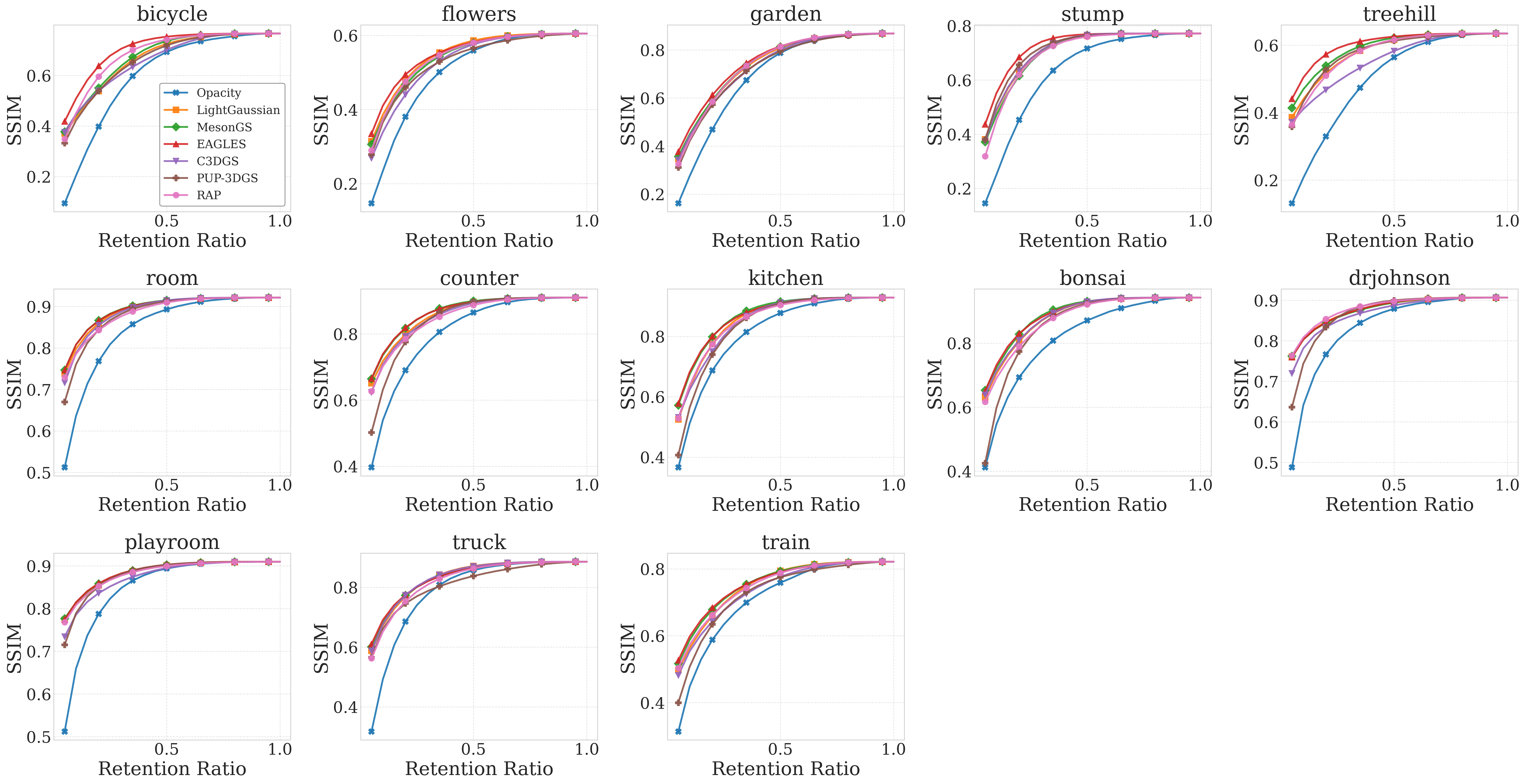}
  \vspace{-3pt}
  \includegraphics[width=\linewidth,height=0.31\textheight,keepaspectratio]{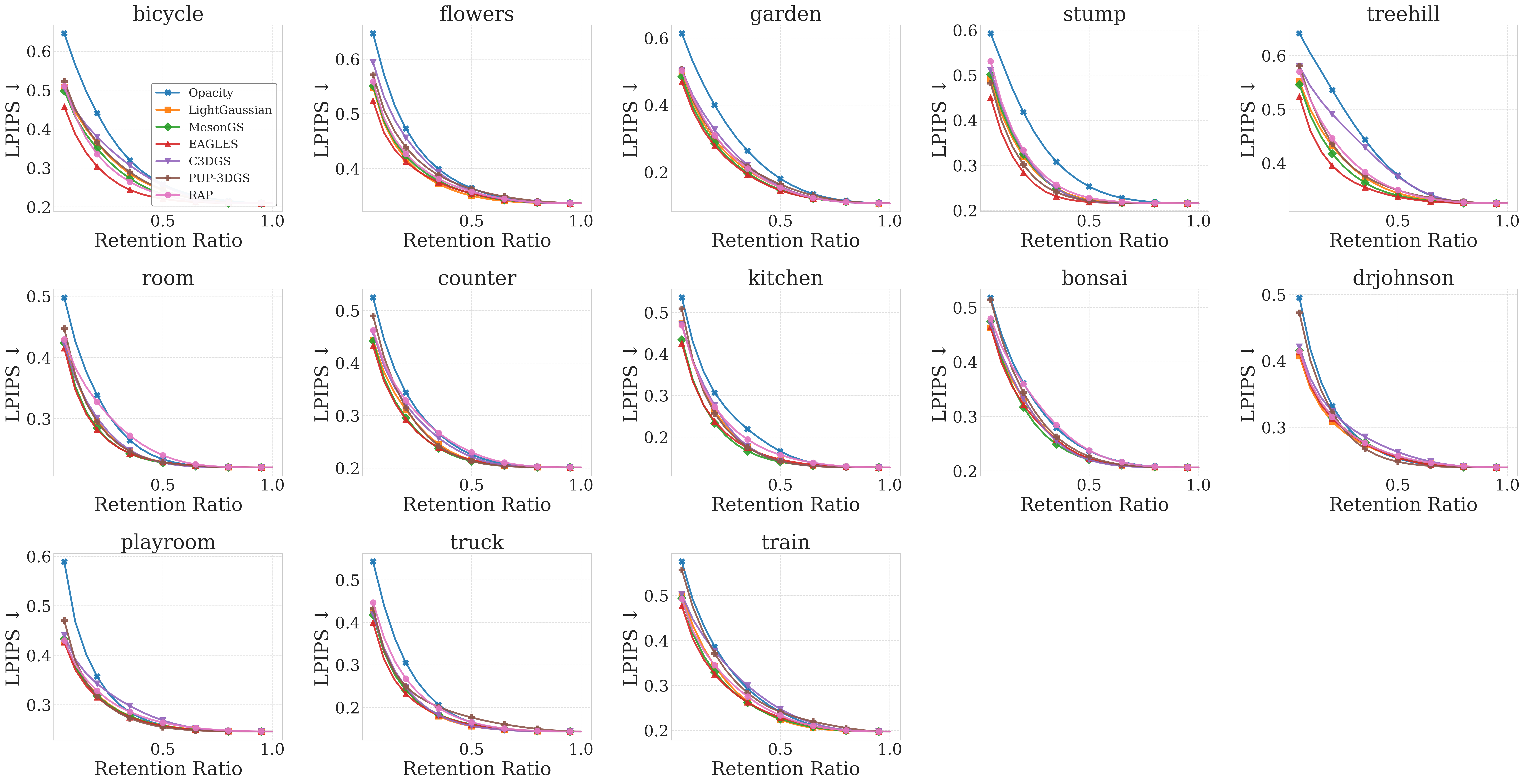}
  \vspace{-6pt}
  \caption{
    Per-scene post-hoc pruning results across three metrics: 
    PSNR (top), SSIM (middle), and LPIPS (bottom).
  }
  \label{fig:sup_2_metrics}
\end{figure*}

Per-scene results are shown in Fig.~\ref{fig:sup_2_metrics}, RAP consistently achieves the highest PSNR on nearly all scenes, with the only exceptions being \textit{bicycle} and \textit{stump} from the Mip-NeRF360 Outdoor dataset, where EAGLES is approximately 0.3\,dB better around the 60\% pruning ratio. 
Visibility-based approaches such as LightGaussian, MesonGS, and EAGLES tend to perform slightly worse than RAP (typically within 0.3\,dB). 
Although these methods differ in their use of opacity, blending opacity, or volume weighting, their overall behavior is largely comparable. Each achieves slightly better results on certain scenes and slightly worse on others, but none exhibits a consistent or systematic advantage across datasets.

In contrast, gradient-based methods (C3DGS, PUP-3DGS) are far less stable. 
They show substantial degradation on several scenes---\textit{kitchen}, \textit{playroom}, \textit{truck}, and \textit{train}---where their PSNR drops by 2--3\,dB relative to RAP. 
At certain pruning ratios, their performance even falls below the naive opacity baseline (e.g., C3DGS on \textit{playroom}, PUP-3DGS on \textit{truck}).

For SSIM and LPIPS, RAP typically ranks second or third on most Mip-NeRF360 scenes, with EAGLES often obtaining the best perceptual scores. 
To better understand these differences, we visualize the retained primitives on the \textit{garden} scene in Fig.~\ref{fig:exp_garden_vis}. 
The comparison reveals distinct selection biases across methods:
\begin{itemize}
    \item \textbf{Visibility-based methods (e.g., EAGLES).}
    These methods strongly favor primitives near the scene center. 
    This arises because: 
    (1) training views surround the central object, causing central primitives to accumulate many projected contributions while background primitives receive far fewer; and 
    (2) projected area decreases with depth, causing distant primitives to appear smaller and thus receive lower scores. 
    Consequently, foreground structures are well preserved, but background regions are severely underrepresented.
    \item \textbf{Gradient-based methods (e.g., C3DGS).}
    These approaches mainly retain high-frequency edges, while smooth surfaces lose most of their support. 
    Under heavy pruning, this leads to incomplete geometry and strong artifacts.
    \item \textbf{RAP (ours).}
    RAP produces a more uniform selection across the entire scene, preserving both foreground and background content. 
    This explains why RAP achieves the strongest PSNR across datasets---PSNR benefits from globally consistent coverage---whereas SSIM and LPIPS, which emphasize structural similarity, may sometimes favor visibility-based techniques.
\end{itemize}






\section{Loss Function Analysis and Score Distribution Visualization}
\label{sup:loss}

\begin{figure}[t]
    \centering

    \begin{subfigure}{0.48\linewidth}
        \centering
        \includegraphics[width=\linewidth]{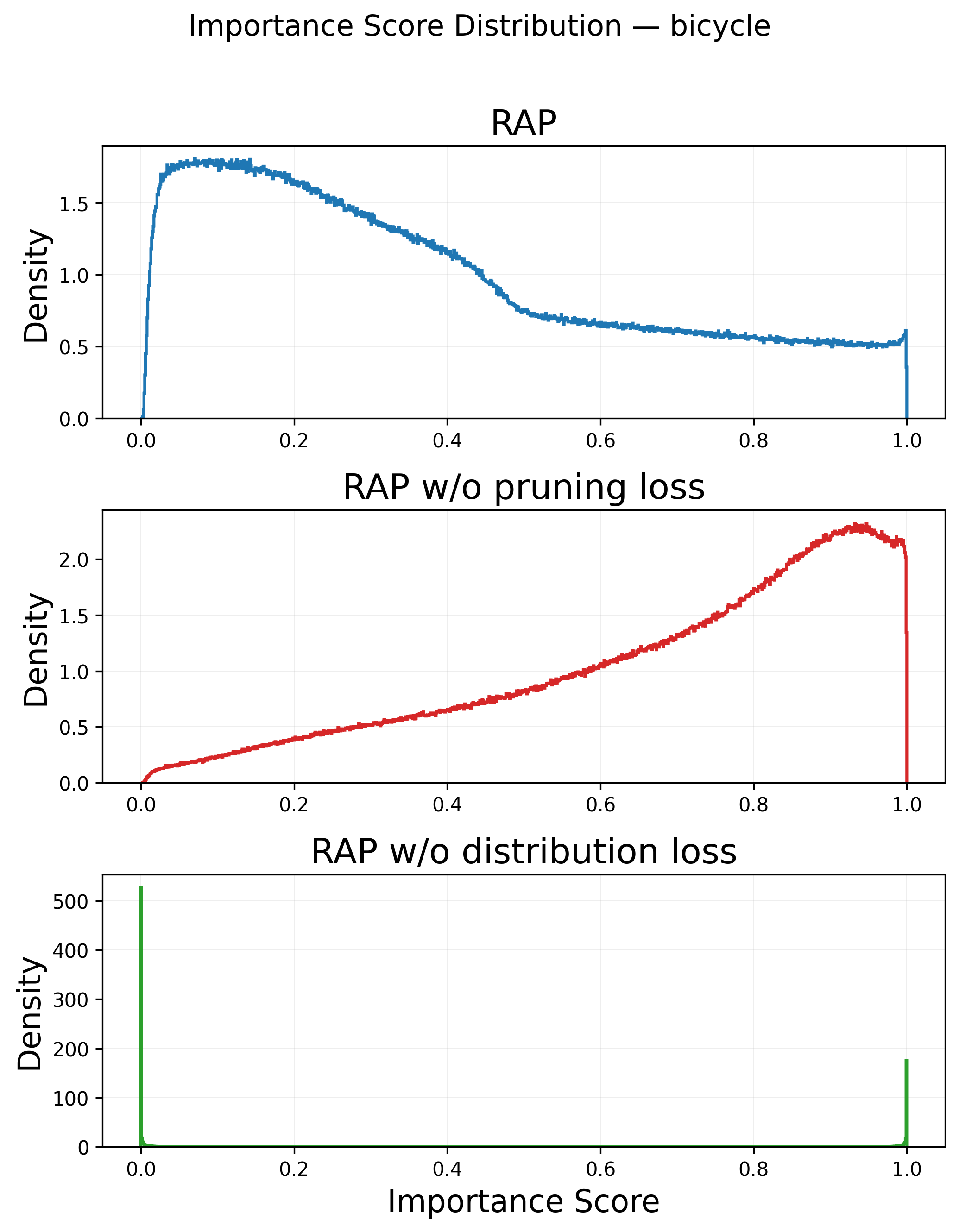}
    \end{subfigure}
    \hfill
    \begin{subfigure}{0.48\linewidth}
        \centering
        \includegraphics[width=\linewidth]{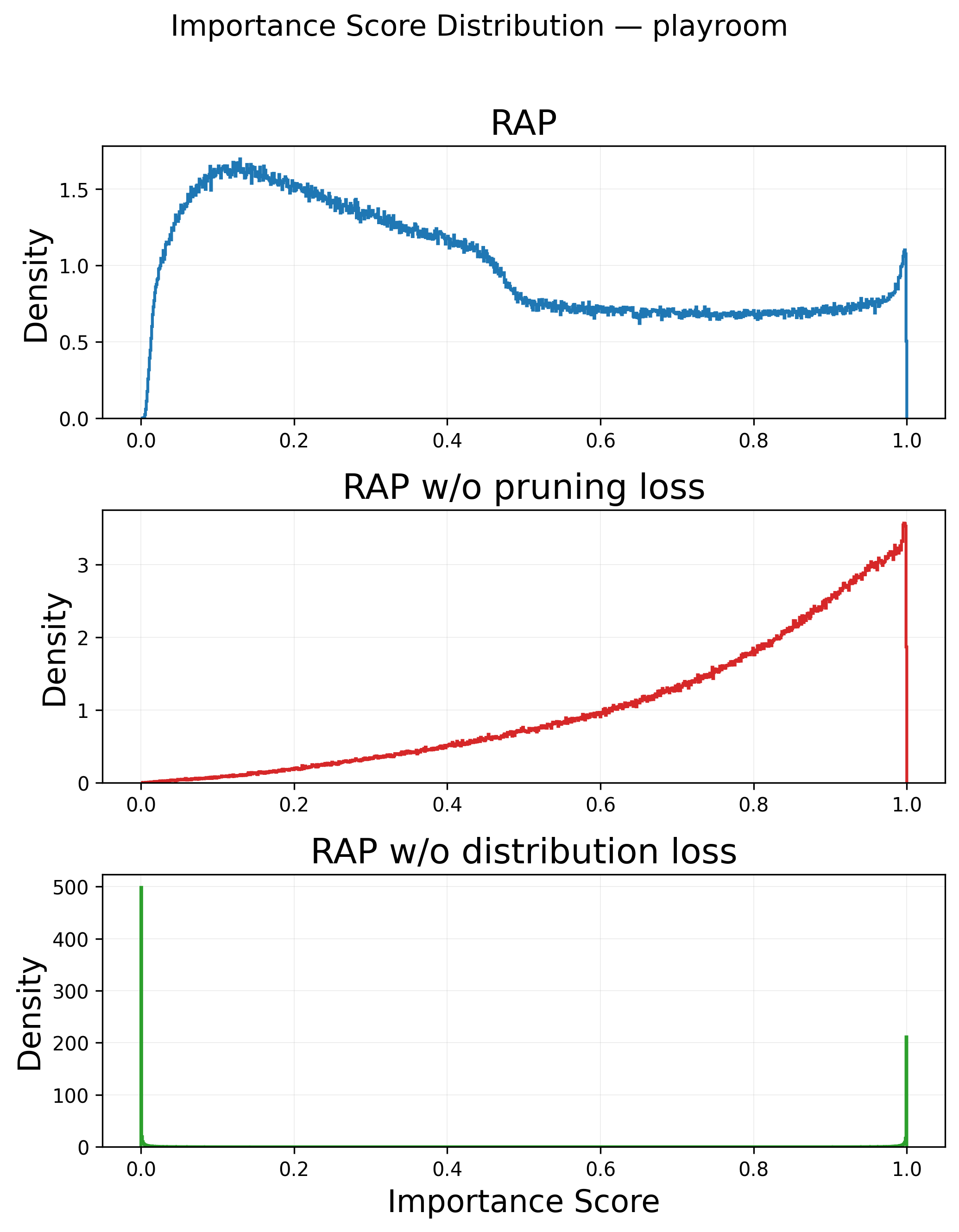}
    \end{subfigure}

    \caption{
    Predicted score distributions for RAP, RAP without pruning-aware loss, 
    and RAP without distribution loss.
    }
    \label{fig:sup_score_distribution}
\end{figure}

To better understand how the pruning-aware loss and the distribution regularization shape the predicted scores, Fig.~\ref{fig:sup_score_distribution} visualizes the score distributions produced by RAP and its ablated variants.
Without the pruning-aware loss, the network tends to assign overly large scores to most primitives, making it difficult to identify truly important points.
In contrast, removing the distribution loss causes the scores to collapse toward near-binary values around 0 and 1, which prevents setting flexible pruning thresholds and makes fine-grained importance discrimination unreliable.
The full RAP model produces a smooth and well-spread distribution—dominated by low scores but with clear separation among mid- and high-importance primitives—enabling stable and accurate pruning across a wide range of ratios.

\end{document}